\documentclass{article}
\usepackage{ijcai21}

\usepackage{times}
\usepackage{soul}
\usepackage{comment}
\usepackage{url}
\usepackage{amssymb}
\usepackage[hidelinks]{hyperref}
\usepackage[utf8]{inputenc}
\usepackage[small]{caption}
\usepackage{graphicx}
\usepackage{amsmath}
\usepackage{amsthm}
\usepackage{booktabs}
\usepackage{algorithm}
\usepackage{algorithmic}
\usepackage{multirow}
\usepackage{subfigure}
\graphicspath{{img/}}

\title{MUTEN: Boosting Gradient-Based Adversarial Attacks via Mutant-Based Ensembles}

\author{
Yuejun Guo \and
Qiang Hu \and
Maxime Cordy \and
Michail Papadakis \and
Yves Le Traon
\affiliations
$^1$Interdisciplinary Centre for Security, Reliability and Trust, University of Luxembourg\\
}

\begin{document}

\maketitle

\begin{abstract}
   Deep Neural Networks (DNNs) are vulnerable to adversarial examples, which causes serious threats to security-critical applications. This motivated much research on providing mechanisms to make models more robust against adversarial attacks. Unfortunately, most of these defenses, such as gradient masking, are easily overcome through different attack means. In this paper, we propose MUTEN, a low-cost method to improve the success rate of well-known attacks against gradient-masking models. Our idea is to apply the attacks on an ensemble model which is built by mutating the original model elements after training. As we found out that mutant diversity is a key factor in improving success rate, we design a greedy algorithm for generating diverse mutants efficiently. Experimental results on MNIST, SVHN, and CIFAR10 show that MUTEN can increase the success rate of four attacks by up to 0.45.

\end{abstract}

\section{Introduction}
\label{sec:intro}
Deep neural networks (DNNs) have achieved impressive success in a wide range of artificial intelligence tasks, such as image classification \cite{Rawat2017deep} and speech recognition \cite{Zhang2018deep}. However, most DNNs suffer from being fooled by adversarial examples, which can result in serious security issues \cite{Finlayson1287,eykholt2018robust}. Such an adversarial example is typically crafted by adding a subtle (elusive) perturbation to a benign input in a way that misleads a DNN model. The adversarial examples are useful not only to reveal security threats in DNNs, but also to improve their robustness, e.g., through adversarial training \cite{madry2018towards}.

Many adversarial attacks have been investigated. In general, there are two types of methods depending on what model information is available. Black-box attacks have no knowledge of the model, while the white-box attacks have access to model information such as its architecture, gradient, and weights. Most adversarial attacks are \emph{gradient-based}, in the sense that they utilize the gradient of the loss function with the hope to compute the perturbation direction that will maximize the likelihood of misclassification. 

\begin{figure}
    \centering
    \subfigure[A model]{
    \label{fig:gradientO}
    \includegraphics[scale=0.295]{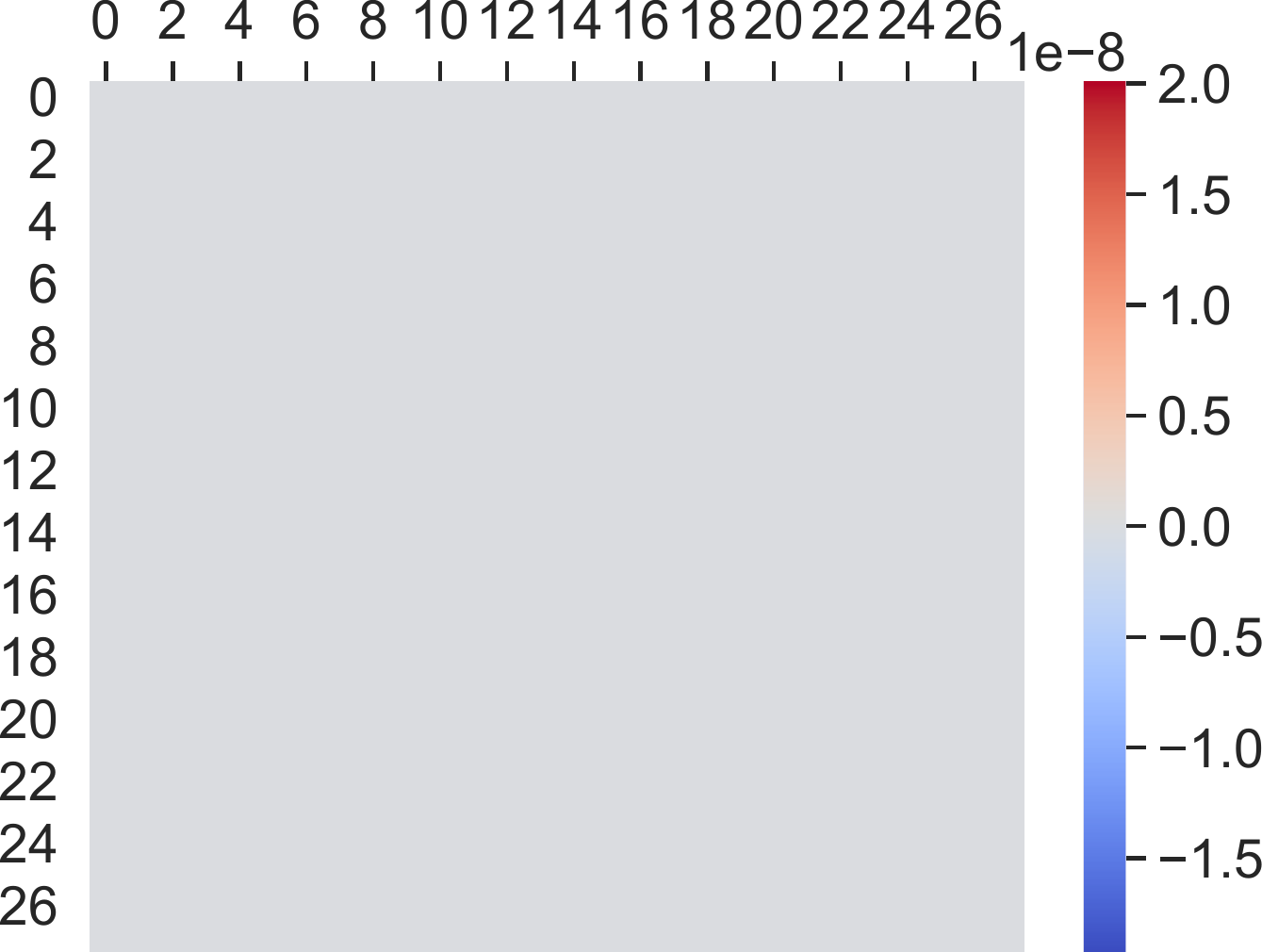}
    }
    \subfigure[An ensemble]{
    \includegraphics[scale=0.295]{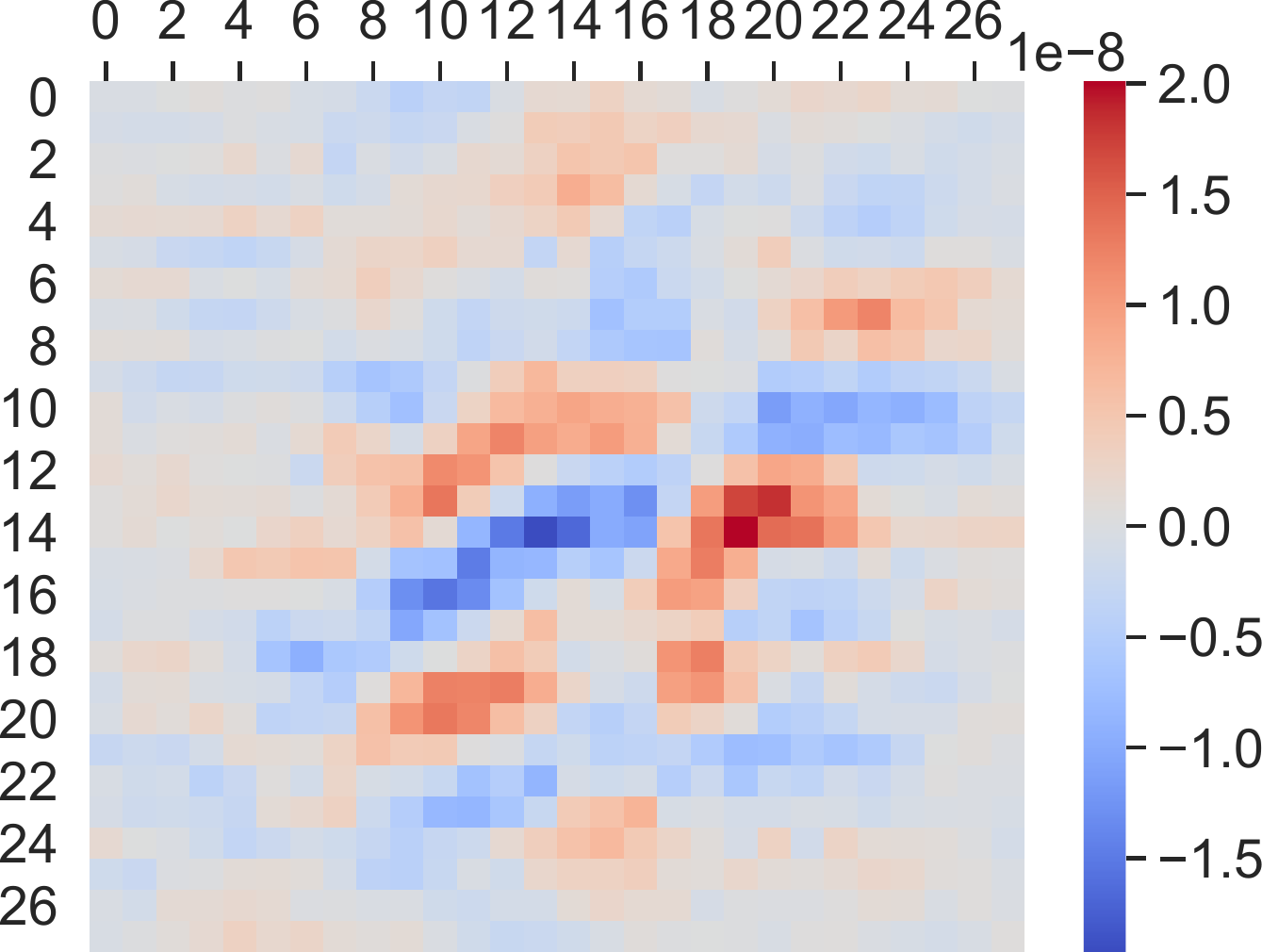}
    \label{fig:gradientE}
    }
    \caption{An illustration of the gradient masking phenomenon and how our approach solves it. Both figures show the gradients of a Lenet-5 model and our produce ensemble with respect to a 28×28 pixel MNIST image (model's output: 92.73\% probability belong to class ``8''). Here, we see that gradient masking occurs in the original model but not in the ensemble.}
    \label{fig:gra}
\end{figure}

The adversarial phenomena, together with the increasing use of machine learning models in real-world applications, have motivated much research on effective ways to generate adversarial examples and defend against them (making the models \emph{robust} to adversarial attacks). Gradient masking \cite{Papernot2017practical} is one such protection means which is commonly used by researchers.\footnote{Back in ICLR 2018, 7 out of 9 papers on adversarial defense were relying on gradient masking. \cite{athalye2018Obfuscated}} It refers to the phenomenon where the gradient cannot be used to optimize the DNN loss function and produce a corresponding perturbation. Gradient masking can result from intentional actions (defenses relying on gradient obfuscation) or natural properties of deep models (where successive computations or specific components cause gradient vanishing); see Figure \ref{fig:gradientO} for an illustration. 

Perhaps paradoxically, despite its popularity the use of gradient masking as a defense mechanism has been fiercely dismissed \cite{athalye2018Obfuscated,Uesato2018risk}. The main reason is that the security it offers is superficial and gives a \emph{``false sense of security''}. For instance, Athalye et al. \cite{athalye2018Obfuscated} show that although gradient masking successfully defends against common gradient-based attacks (e.g. PGD \cite{madry2018towards} and C\&W \cite{cw2017}), it is easily overcome through alternative means (e.g., approximated derivatives, Expectation Over Transformation \cite{Athalye2018synthesizing} and reparameterization).

In this paper, we further reduce the trust that one can have in this defense mechanism and show that \emph{common attacks can easily overcome gradient masking}. We propose MUTEN, a fast and effective way to attack models with masked gradients. Unlike previous work circumventing gradient masking \cite{athalye2018Obfuscated,Uesato2018risk}, MUTEN does not rely on new attack means and can be applied jointly with any existing algorithms (we consider, more specifically, FGSM \cite{goodfellow2014explaining}, BIM \cite{kurakin2016adversarial}, PGD \cite{madry2018towards} and C\&W \cite{cw2017}). Thereby, our work strengthens the claims that the robustness promised by gradient masking is a mere illusion.

The effectiveness of our approach leans on Liu et al.'s work \cite{Liu2017DelvingAttacks}, where they show that adversarial examples crafted from an ensemble of surrogate models transfer relatively well to the original model. Unlike their method, though, MUTEN avoids the prohibitive cost of training multiple models. More precisely, we generate an ensemble of models containing a set of \emph{mutant} models, obtained by altering the original one \emph{after training} (e.g., by introducing random noise into its weights). Working in white-box settings, we also include the original model in the ensemble. Thus, the overhead of attacking the ensemble is limited to applying the attack to each mutant and computing the average of the perturbations. 

Given that there exist many ways to mutate a model \cite{ma2018deep}, we investigate which combinations of mutation are more effective in increasing the success rate of the attacks. Our hypothesis is that an effective ensemble should contain \emph{a diverse set of accurate mutants}. On the one hand, we ensure that the mutants retain the performance (test accuracy) of their original model by controlling the proportion of modified weights, neurons, or layers. On the other hand, we propose a greedy algorithm to produce a \emph{diverse} mutants where the diversity is measured by the centered kernel alignment (CKA) metric \cite{kornblith2019similarity} and PageRank algorithm \cite{moler2011experiments}. 

To summarize, the main contributions of this paper are:
\begin{enumerate}
    \item We propose MUTEN, a novel approach to boost the effectiveness of gradient-based attacks, overcoming gradient masking, by building an ensemble from the original model without extra training.
    \item We plug MUTEN on top of existing adversarial attacks without requiring modifications to their core algorithms.
    \item We design a greedy algorithm to generate diverse mutants that retain the original model accuracy.
    \item We show the effectiveness of our approach through an extensive experimental study involving three datasets, four models, and four attacks. Our results reveal that the success rate can be increased by up to 0.45.
\end{enumerate}

\section{Background}
\label{sec:back}
\subsection{Gradient-Based Attacks}
\label{sec:background-attack}

\paragraph{FGSM} Goodfellow \emph{et al.} \cite{goodfellow2014explaining} proposed one of the simplest methods, fast gradient sign method (FGSM), to generate adversarial examples. Let $x$ and $x'$ be the original image and the adversarial example, respectively, FGSM crafts $x'$ by
$
    x'=x+\epsilon\cdot{\rm sign}\left(\triangledown_{x}J\left(x,y\right)\right)
$
where $\epsilon$ controls the perturbation size. $sign\left(\cdot\right)$ is the sign function. The sign of a real number is -1 for a negative value, 1 for a positive value, and 0 for value 0. $\triangledown_{x}J\left(x,y\right)$ computes the gradient of the training loss $J$ given $x$ and its true class $y$.

\paragraph{BIM} Based on FGSM, Kurakin \emph{et al.} \cite{kurakin2016adversarial} proposed an iterative version, basic iterative method (BIM, also called i-FGSM), which applies FGSM multiple times with a small step size to craft $x'$:
$
    x'_{n+1}=Clip_{x,\epsilon}\left\{x'_{n}+\alpha\cdot{\rm sign}\left(\triangledown_{x}J\left(x'_{n},y\right)\right)\right\}
$
where $x'_0=x$ and $Clip_{x,\epsilon}$ is a clip function applied after each iteration to ensure that the result is still in $\epsilon$-neighbourhood of $x$. $n$ is the iteration index and $\alpha$ is the step size.

\paragraph{PGD} Madry \emph{et al.} \cite{madry2018towards} proposed the projected gradient descent (PGD) to improve BIM. The difference is that BIM starts from the original point, while PGD randomly chooses the starting point within a $\epsilon$ norm ball.

\paragraph{C\&W} The C\&W attack, proposed by Carlini and Wagner \cite{cw2017}, is known as one of the strongest attacks to DNNs. Instead of using the training loss, C\&W uses a designed loss function $f$ to craft the adversarial example $x'$ which minimizes
$
    D\left(x,x'\right)+c\cdot f(x')
$ 
where $D$ is a distance metric and $c$ is a constant that controls the distance and the confidence of $x'$.

\subsection{Mutation of Deep Learning Models}
Post-training mutation of DNNs has been applied mainly for quality assurance purposes. Due to the unique characters of DNN models, various mutation operators have been proposed at source level (modifying the training data or the training program) or at model level \cite{ma2018deep}.

In this paper, we apply model-level mutations, where a mutant is created directly by changing the neurons, weights or, layers slightly without training. In general, modifying the layers requires specific architectures of the DNN models and degrades the performance (accuracy) significantly, and is less applicable. Both the weight- and neuron-level operators work efficiently to generate mutants and are more widely used.

Recent studies have shown the utility of mutation in different tasks. Ma \emph{et al.} \cite{ma2018deep} propose to mutate test data class by class to figure out the weakness in test data, which is helpful to check to bias in data. Hu \emph{et al.} \cite{Hu2019ddep} points out that by a defined killing score metric, the mutants can be used to validate how robust a DNN model is against an input data or its segment. In \cite{Wang2019adver}, Wang \emph{et al.} assume that the adversarial examples are near the decision boundary, thus, the data that change the labels by different mutants are considered as adversarial.

\section{Approach}
\label{sec:approach}
We aim to improve the success rate of gradient-based adversarial attacks applied to gradient-masking models. The main idea of MUTEN is to produce a collection of diverse mutant models to build an ensemble, and attack the ensemble instead of the single original model.

\subsection{Diverse Mutant Generation}
\label{subsec:greedy}
To build an ensemble without extra training, we employ the model-level mutation. Since previous research has shown that mutating layers always degrade the performance (test accuracy) significantly \cite{ma2018deep}, we consider 5 operators working only at the weight- and neuron-level. Gaussian fuzzing (GF) operator adds noise to the selected weights following the Gaussian distribution. The weight shuffling (WS) rearranges the selected weights. The neuron effect blocking (NEB) resets the connection weight of a selected neuron to the next layer to zero. The neuron activation inverse (NAI) operator inverts the activation status of a neuron. The neuron switch (NS) exchanges two neurons within the same layer. The mutation ratio controls how much percentage of weights or neurons are selected for mutation in each layer. As applying more mutations has a higher likelihood to decrease model performance, we randomly set it between 1\% and 4\% following the previous findings of Ma et al. \cite{ma2018deep}.

To measure mutant diversity, we use the centered kernel alignment (CKA) \cite{kornblith2019similarity} metric and the PageRank algorithm \cite{moler2011experiments}. More precisely, CKA measures the similarity between DNN representations. Given the input data X, let $H_1$ and $H_2$ be two feature matrices of $X$ by two models, respectively. $H_1$ and $H_2$ are considered as the DNN representations. The similarity between $H_1$ and $H_2$ is defined by

\begin{equation}
    CKA\left(K,L\right)=\frac{{\rm HSIC}\left(K,L\right)}{\sqrt{{\rm HSIC}\left(K,K\right){\rm HSIC}\left(L,L\right)}}
\end{equation}
where $K$ and $L$ are the kernel matrices by passing $H_1$ and $H_2$ through kernels. $\rm HSIC$ is the Hilbert-Schmidt independence criterion. Like \cite{Chen2020pace}, we use the output of the last hidden layer in a DNN as the feature. 

The PageRank algorithm aims at measuring the importance/rank of website pages where a page linked to by many pages has a high rank. Inspired by this, taking a mutant as a website page and the similarity as the linking weight, we assume that the mutant with a low rank is diverse within the mutant set as it is dissimilar with the others. 

\begin{figure}
    \centering
    \subfigure[Diverse]{
    \includegraphics[scale=0.28]{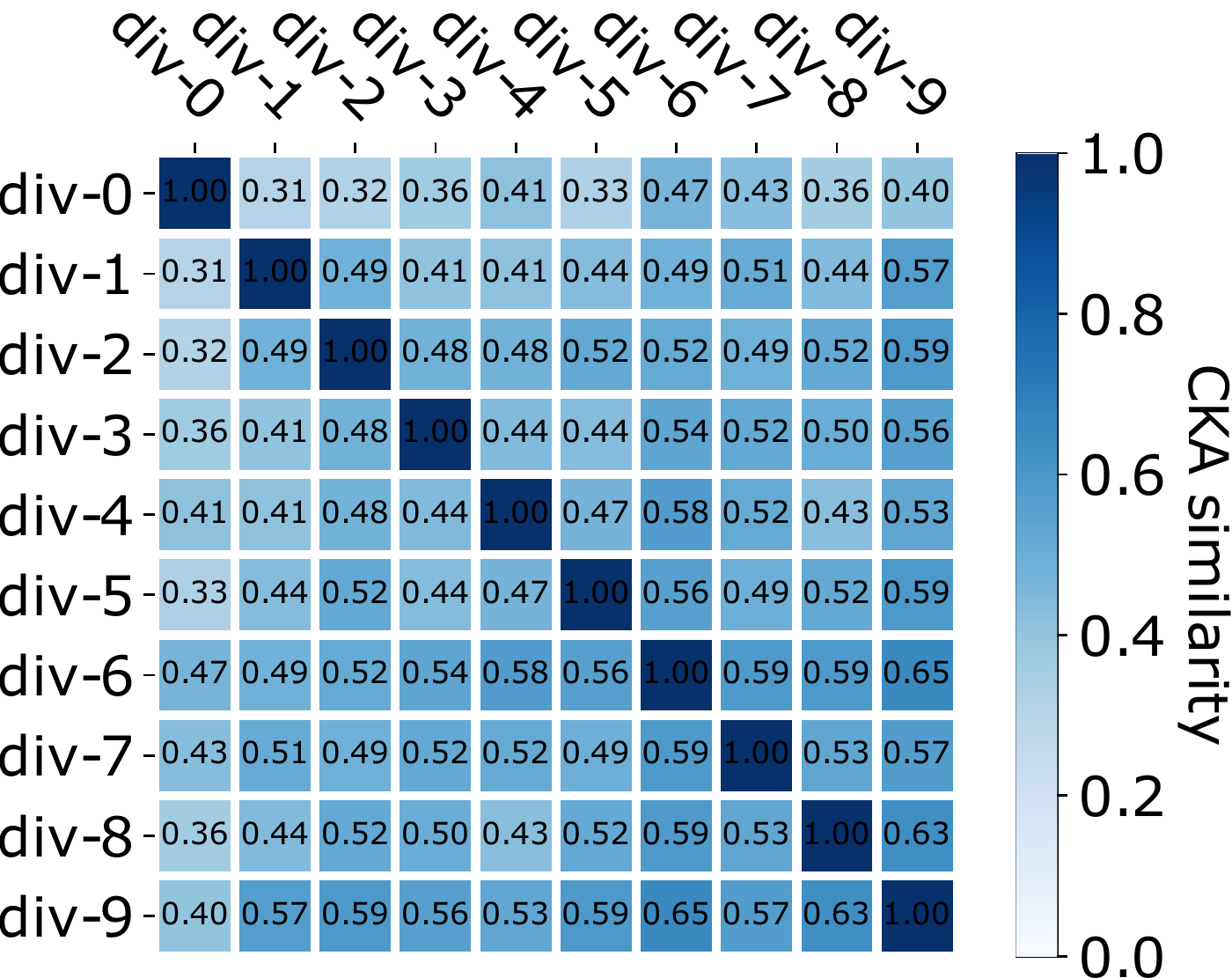}
    }\\
    \vspace{-1.5mm}
    \subfigure[Random]{
    \includegraphics[scale=0.28]{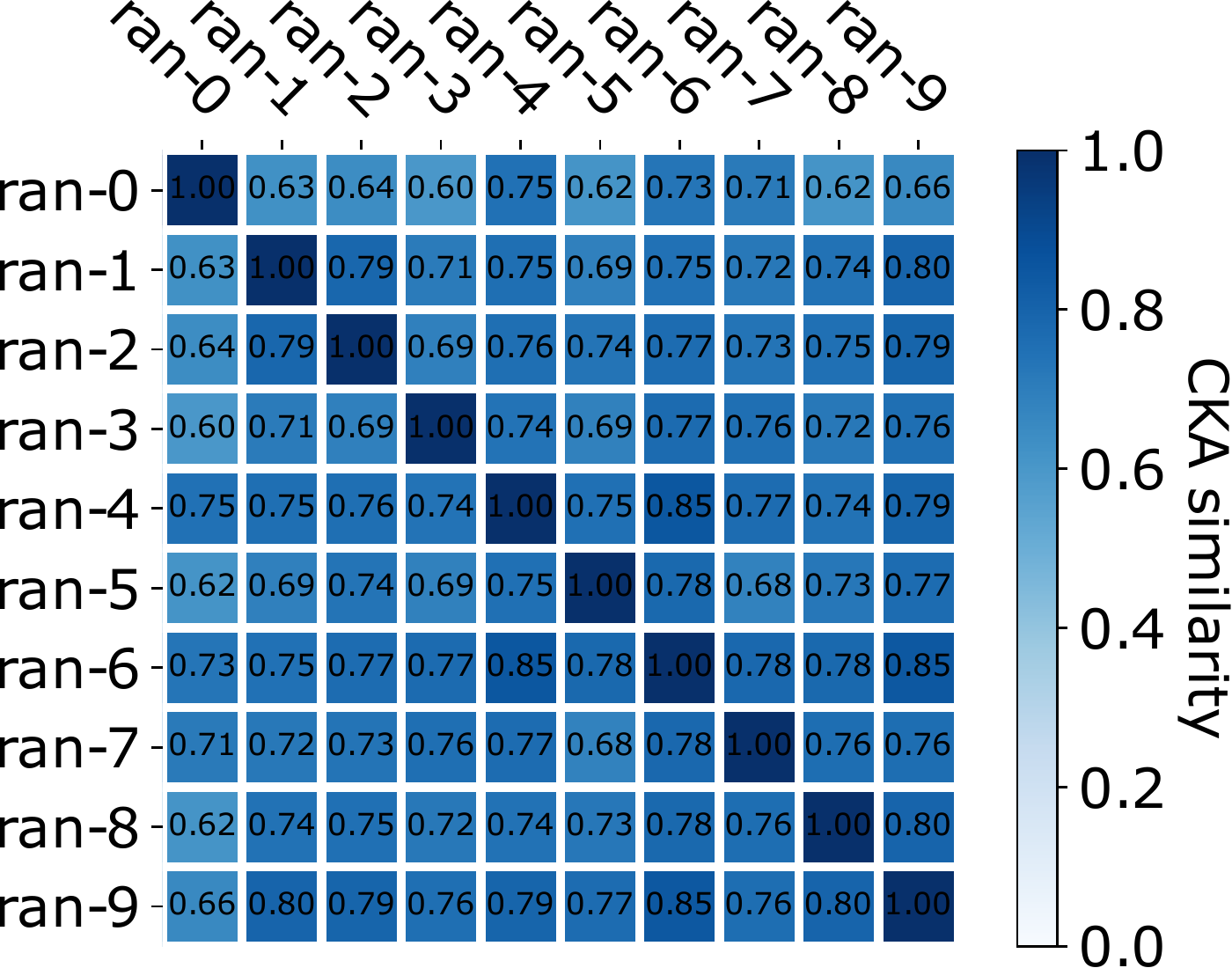}
    }
    \subfigure[Similar]{
    \includegraphics[scale=0.28]{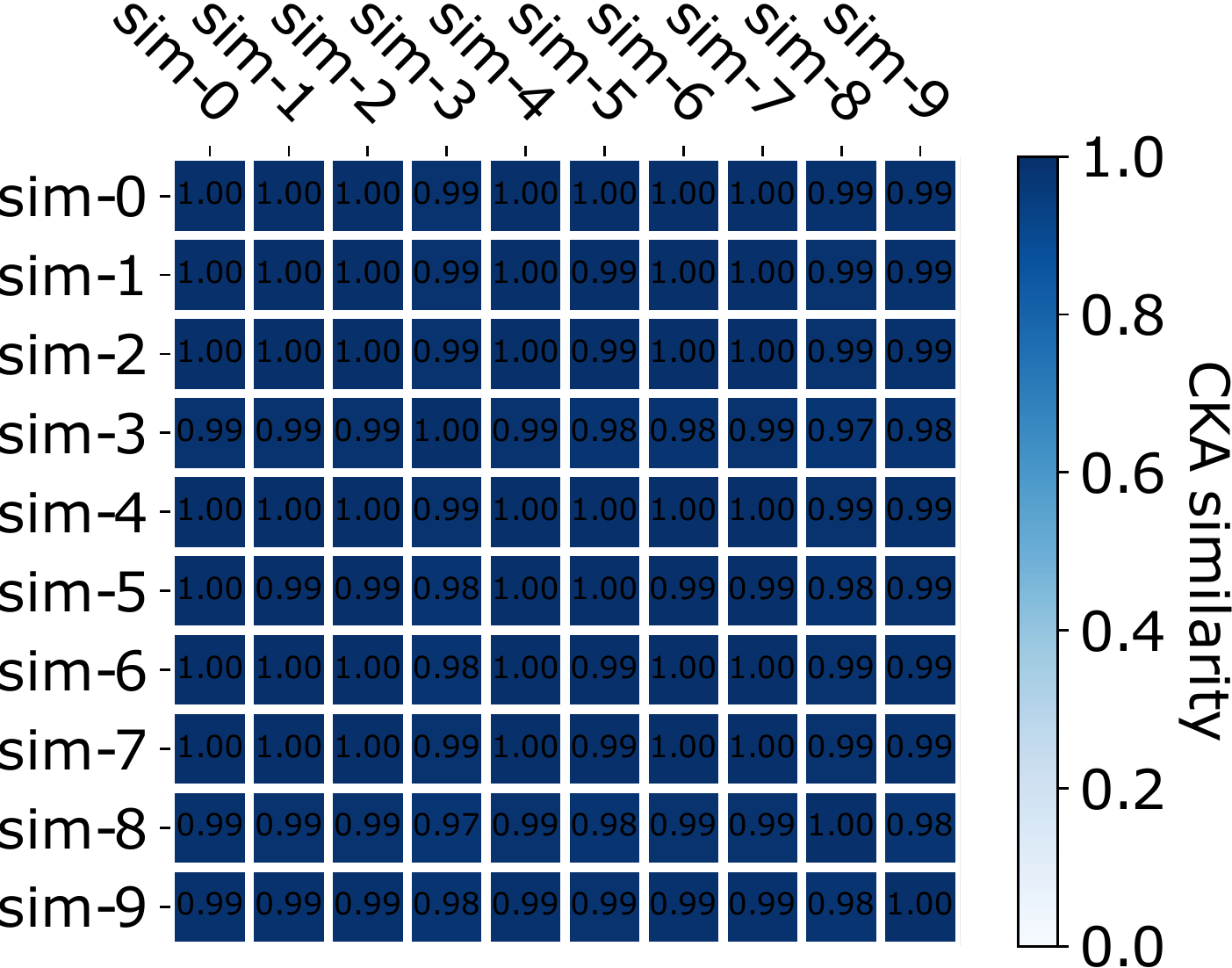}
    }
    \caption{CKA similarity of 10 mutants (diverse, random and similar). Darker color indicates a higher similarity between models. Dataset: MNIST; Model: Lenet-5.}
    \label{fig:distance}
\end{figure}

Algorithm \ref{alg:greedy} shows our overall generation method. To increase the diversity of generated mutants, we use 20 pairs of mutation operators and ratios. Given a model $M$ and a required number of mutants $n$, the mutant set $set$ and a similarity matrix $D$ are initialized to be empty, and a counting index $count$ is used to control the termination of the algorithm (Line 1). In each iteration, a pair of mutation operator and ratio is randomly selected from all the candidates (Line 3) to generate a mutant (Line 4). After the first iteration, the mutant set $Mu$ is updated with a mutant (Lines 5-6). When another new mutant is generated, first, we compute the linear CKA similarity between this mutant and the ones in $Mu$ to update the similarity matrix $D$ (Lines 7-9). If the size of mutation set, $|Mu|$, is smaller than $n$, the procedure continues, otherwise, $D$ is fed into the $pageRank$ function to compute the diversity and update the mutant set (Lines 11-13 ). The maximum size of $Mu$ is $n$. The iteration terminates until it reaches a preset number of iterations. Figure \ref{fig:distance} shows the linear CKA similarity of 10 mutants in types of diverse, random, and similar, respectively. Diverse mutants are more dissimilar from each other.

\begin{algorithm}
\caption{Greedy mutant generation algorithm}
\label{alg:greedy}
\textbf{Input}: $M$: DNN model\\
\hspace*{1cm} $O=\{$GF, WS, NEB, NAI, NS$\}$: mutation operators\\
\hspace*{1cm} $R=\left\{0.01, 0.02, 0.03, 0.04\right\}$: mutation ratios\\
\hspace*{1cm} $n$: required number of mutants\\
\hspace*{1cm} $ite$: number of iterations\\
\textbf{Output}: $Mu$: a set of mutants
\begin{algorithmic}[1]
\STATE Initialize $Mu$, $D$, $count=0$
\WHILE{$count<ite$}
\STATE $\left(o,r\right)=randomSelect\left(O,R\right)$
\STATE $m=mutantGenerator\left(M,o,r\right)$
\IF {$|Mu|$==0}
\STATE $Mu=\left\{m\right\}$
\ELSE
\STATE $D=computeCka\left(m, Mu\right)$
\IF {$|Mu|<n$}
\STATE $|Mu|=Mu\cup \left\{m\right\}$
\ELSE
\STATE $Mu=pageRank\left(D, Mu, m\right)$
\ENDIF
\ENDIF
\STATE $count++$
\ENDWHILE
\STATE \textbf{return} $Mu$
\end{algorithmic}
\end{algorithm}

\subsection{Ensemble Model Construction}
\begin{figure}[h]
    \centering
    \includegraphics[width=0.5\textwidth]{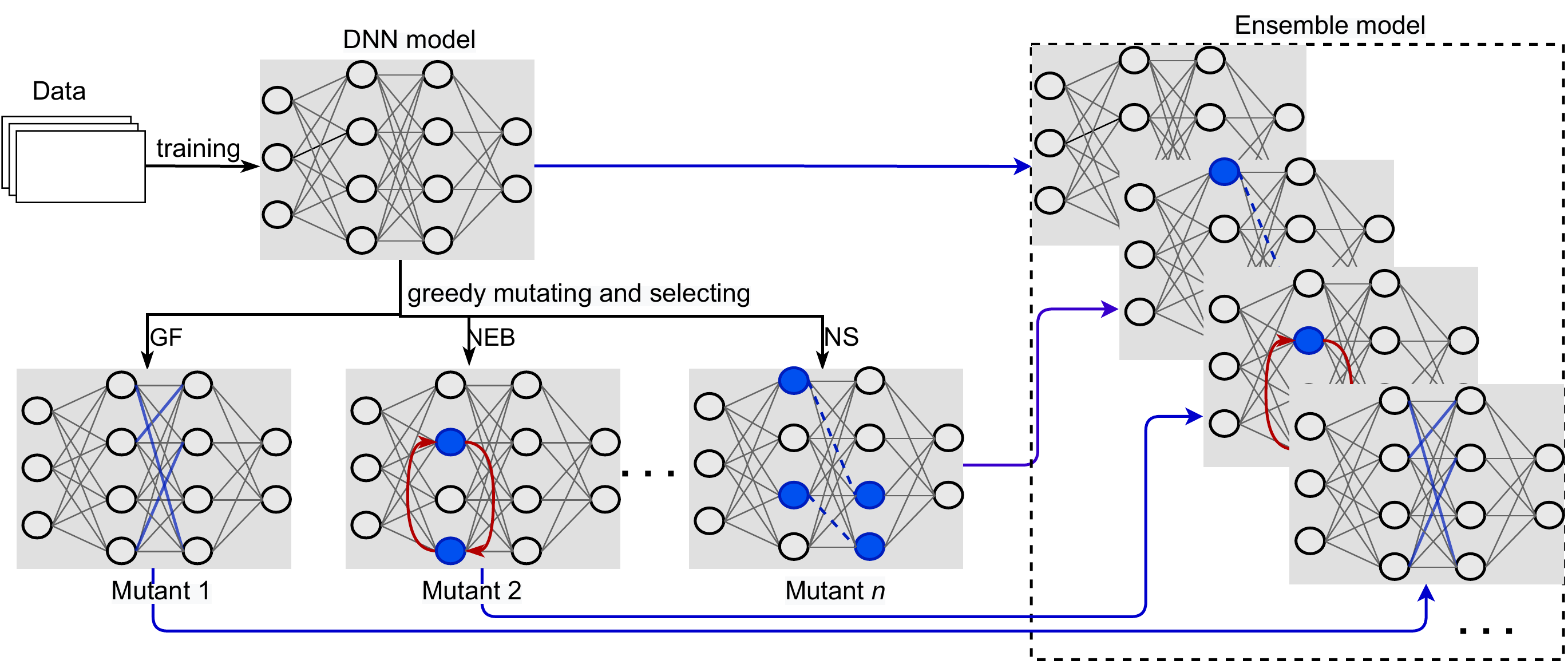}
    \caption{Example of building an ensemble model.}
    \label{fig:ensemble}
\end{figure}

Figure \ref{fig:ensemble} illustrates how we construct an ensemble. Given the training data, a model is trained with a specific architecture and parameters. By the greedy algorithm mentioned in Section \ref{subsec:greedy}, multiple diverse mutants are obtained. In the example, we show the effect of the mutation operators GF, NEB, and NS, and the modified weights and neurons are highlighted in blue. At last, an ensemble model is built by gathering all the original model and its mutants. When performing an adversarial attack, we use the simple average strategy \cite{demir2016ensemble}. That is, the attack accesses each base model to obtain the gradient given an input data, then the perturbation is calculated based on the average of all the gradients.

\section{Experiments}
\label{sec:exp}
To evaluate the effectiveness of our approach in increasing adversarial attack success rate, we conducted experiments using TensorFlow 2.3.0 and Keras 2.4.3. We implemented MUTEN by extending the DeepMutation++ tool \cite{Hu2019ddep} and apply the library of CKA\footnote{\url{https://github.com/google-research/google-research/tree/master/representation_similarity}} and fast-pagerank\footnote{\url{https://github.com/asajadi/fast-pagerank}} to compute the rank of diversity. To allow for reproducibility, our full implementation and evaluation subjects are available on GitHub.\footnote{Link anomymized for double-blind review.}

\subsection{Experimental Setup}
\paragraph{DNN models and datasets} We conduct the experiments on 3 widely used datasets, MNIST \cite{Lecun1998gradient}, SVHN \cite{Netzer2011reading}, and CIFAR10 \cite{Alex2009techm}. MNIST is a 10-class grayscale handwritten digit dataset. SVHN is a real-world image dataset including 10 classes of street view house numbers. CIFAR10 is a 10-class dataset with color images. For MNIST and SVHN, we employ the Lenet-5 model as \cite{athalye2018Obfuscated}. For CIFAR10, we use VGG16 and ResNet20V1 because, as deep models, they are subject to natural gradient vanishing phenomena. Table \ref{tab:datasets} summarizes the detailed information of the datasets and models. As we generate weight- and neuron-level mutants, the number of weights and neurons are also given in the third and fourth columns, respectively.

\begin{table}
\centering
\resizebox{0.5\textwidth}{!}{
\begin{tabular}{lrrrrr}
\toprule
Dataset & Model & \#Weights & \#Neurons & \#Tests & Accuracy(\%) \\
\midrule
MNIST & Lenet-5 & 107550 & 236 & 10000 & 98.89 \\
SVHN & Lenet-5 & 136650 & 236 & 26032 & 88.93 \\
CIFAR10 & ResNet20V1 & 270896 & 794 & 10000 & 90.71 \\
 & VGG16 & 2851008 & 1674 & 10000 & 91.17 \\
\bottomrule
\end{tabular}}
\caption{DNN models and datasets}
\label{tab:datasets}
\end{table}

\paragraph{Attacks and parameter setting} We evaluate MUTEN with four state-of-the-art gradient-based attacks (see Section \ref{sec:back}), \emph{i.e.}, FGSM, BIM, PGD ($l_\infty$-norm), C\&W ($l_2$-norm). The attacks are implemented based on the IBM ART framework\footnote{\url{https://github.com/Trusted-AI/adversarial-robustness-toolbox}} \cite{art2018}. To avoid the influence of parameters, we set 7 values for each attack. Table \ref{tab:attacks} presents the configurations of the attacks used for each dataset. Besides, in FGSM, BIM, PGD, the step size of $\epsilon$ is set as $\frac{\epsilon}{10}$, and the maximum iteration is set as 40 and 20, for MNIST/SVHN and CIFAR10, respectively. In C\&W, the learning rate is 0.1, and the maximum iteration is 100 for all the models. For the rest of the parameters, we use the default setting in ART. 

\begin{table}
\centering
\resizebox{0.5\textwidth}{!}{
\begin{tabular}{lrr}
\toprule
Dataset & \begin{tabular}[c]{@{}r@{}}FGSM, BIM, PGD\\ $\epsilon$\end{tabular} & \begin{tabular}[c]{@{}r@{}}C\&W\\ $c$\end{tabular} \\
\midrule
MNIST & \multirow{2}{*}{$0.1,0.15,0.2,0.25,0.3,0.35,0.4$} & \multirow{2}{*}{$7,8,9,10,11,12,13$} \\
SVHN &  &  \\
CIFAR10 & $\frac{2}{255}, \frac{4}{255}, \frac{6}{255}, \frac{8}{255}, \frac{10}{255}, \frac{12}{255}, \frac{14}{255}$ & $0.1,0.15,0.2,0.25,0.3,0.35,0.4$ \\
\bottomrule
\end{tabular}
}
\caption{Attack configurations. ``$\epsilon$'' and $c$ stand for the maximum perturbation and initial constant, respectively.}
\label{tab:attacks}
\end{table}

\subsection{Results}
We conducted three series of experiments to evaluate MUTEN from different perspectives. Since it is meaningless to attack misclassified data as being considered as ``adversarial'' for the model, each attack crafts one example based on each benign data. Each experiment was repeated five times to reduce the influence of randomness in both the creation of mutants and the application of the attacks. Reported results are the average of those five runs. 

\begin{figure*}[h]
    \centering
    \subfigure[MNIST, FGSM]{
    \includegraphics[scale=0.26]{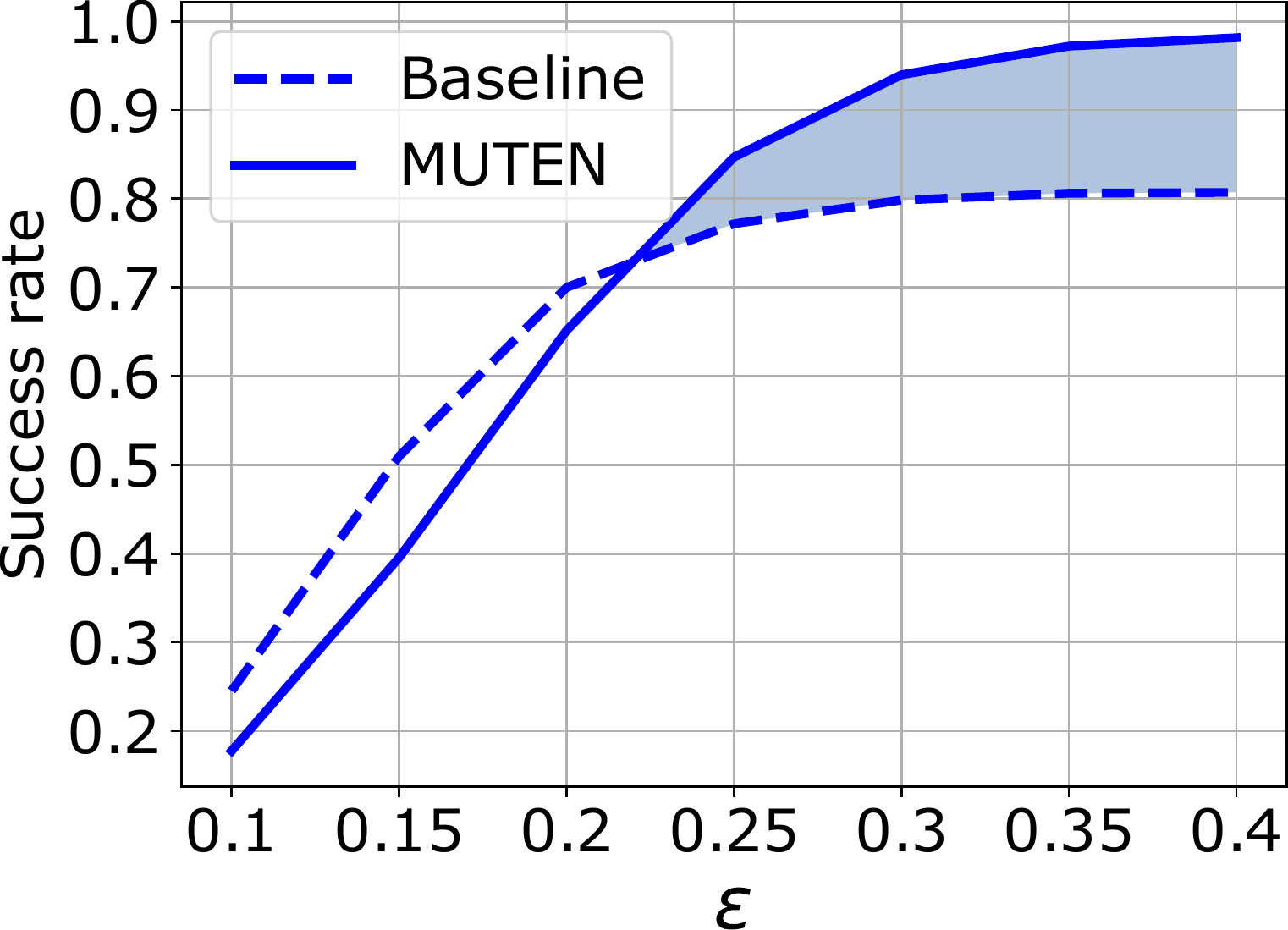}%
    }
    \subfigure[SVHN, FGSM]{
    \includegraphics[scale=0.26]{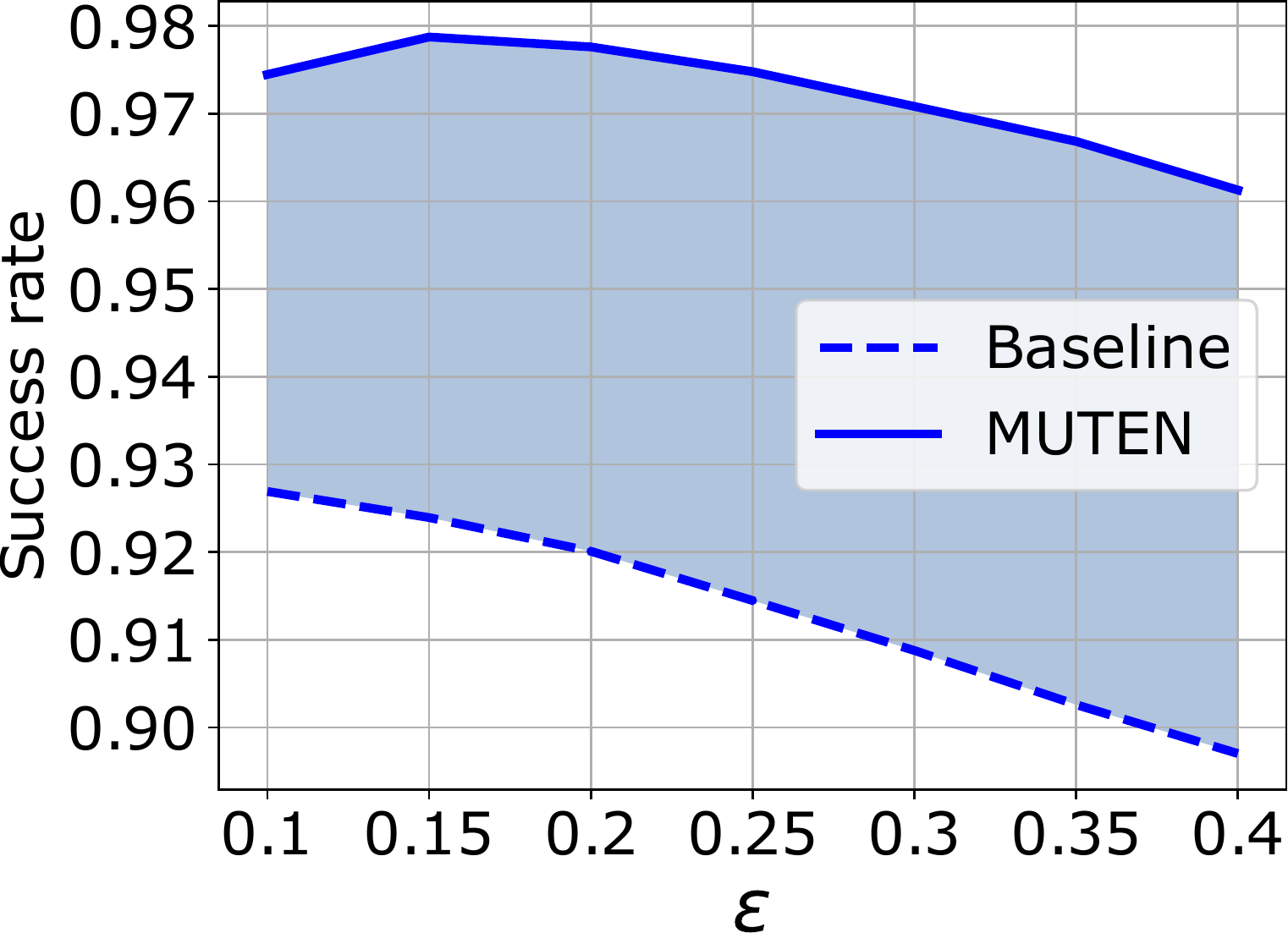}%
    }
    \subfigure[ResNet20V1, FGSM]{
    \includegraphics[scale=0.26]{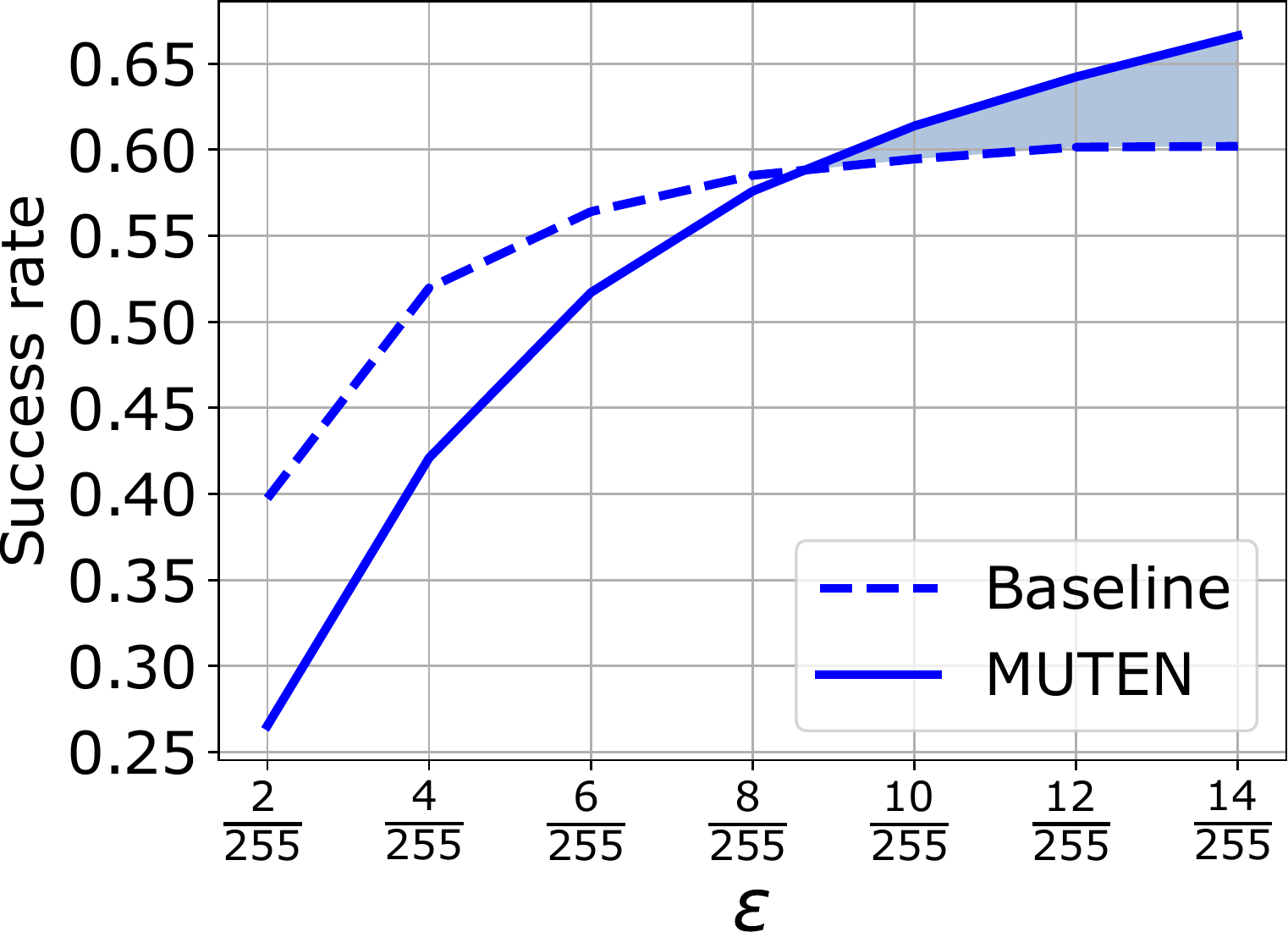}%
    }
    \subfigure[VGG16, FGSM]{
    \includegraphics[scale=0.26]{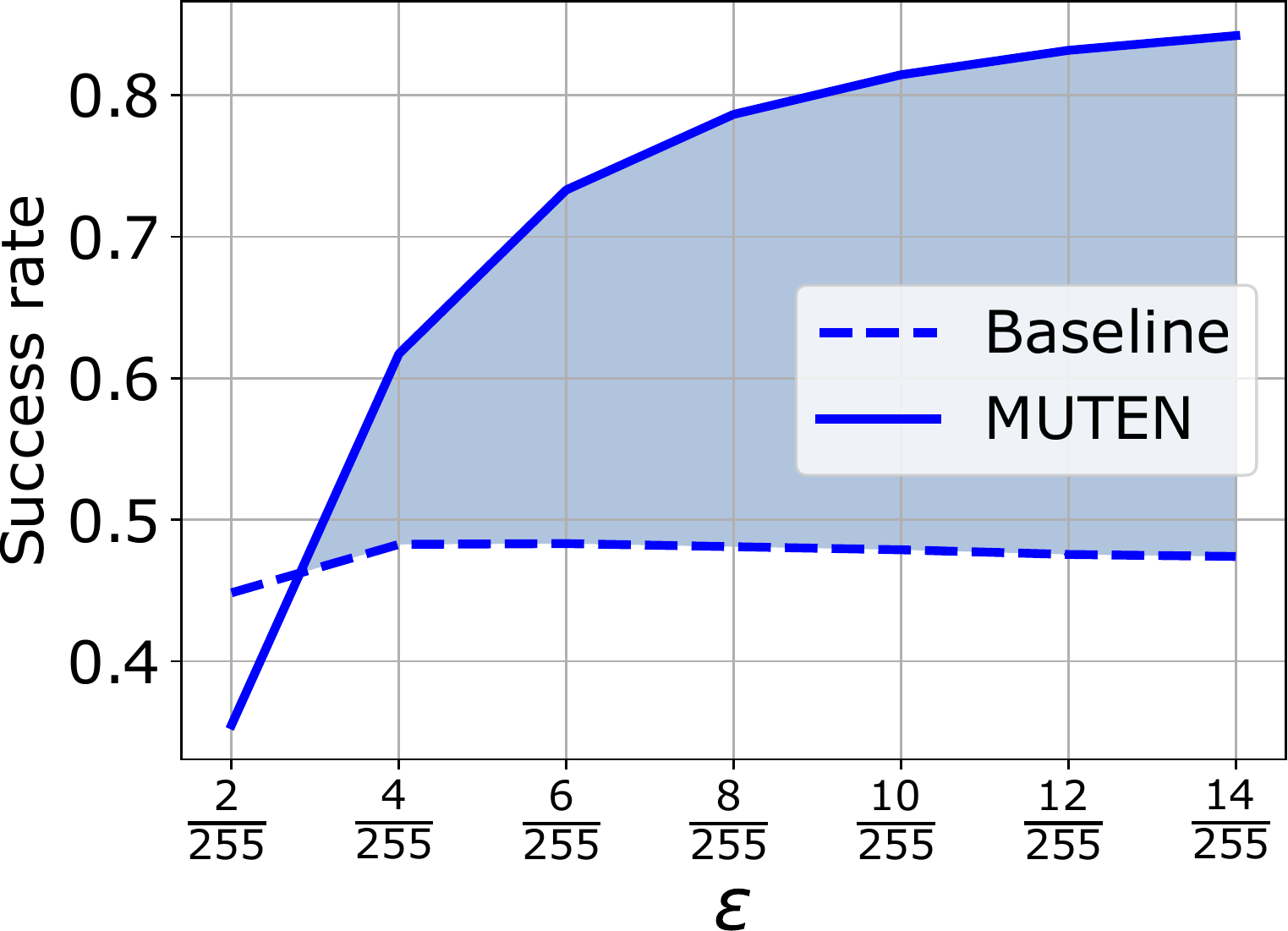}%
    }\\
    \vspace{-1.5mm}
    \subfigure[MNIST, BIM]{
    \includegraphics[scale=0.26]{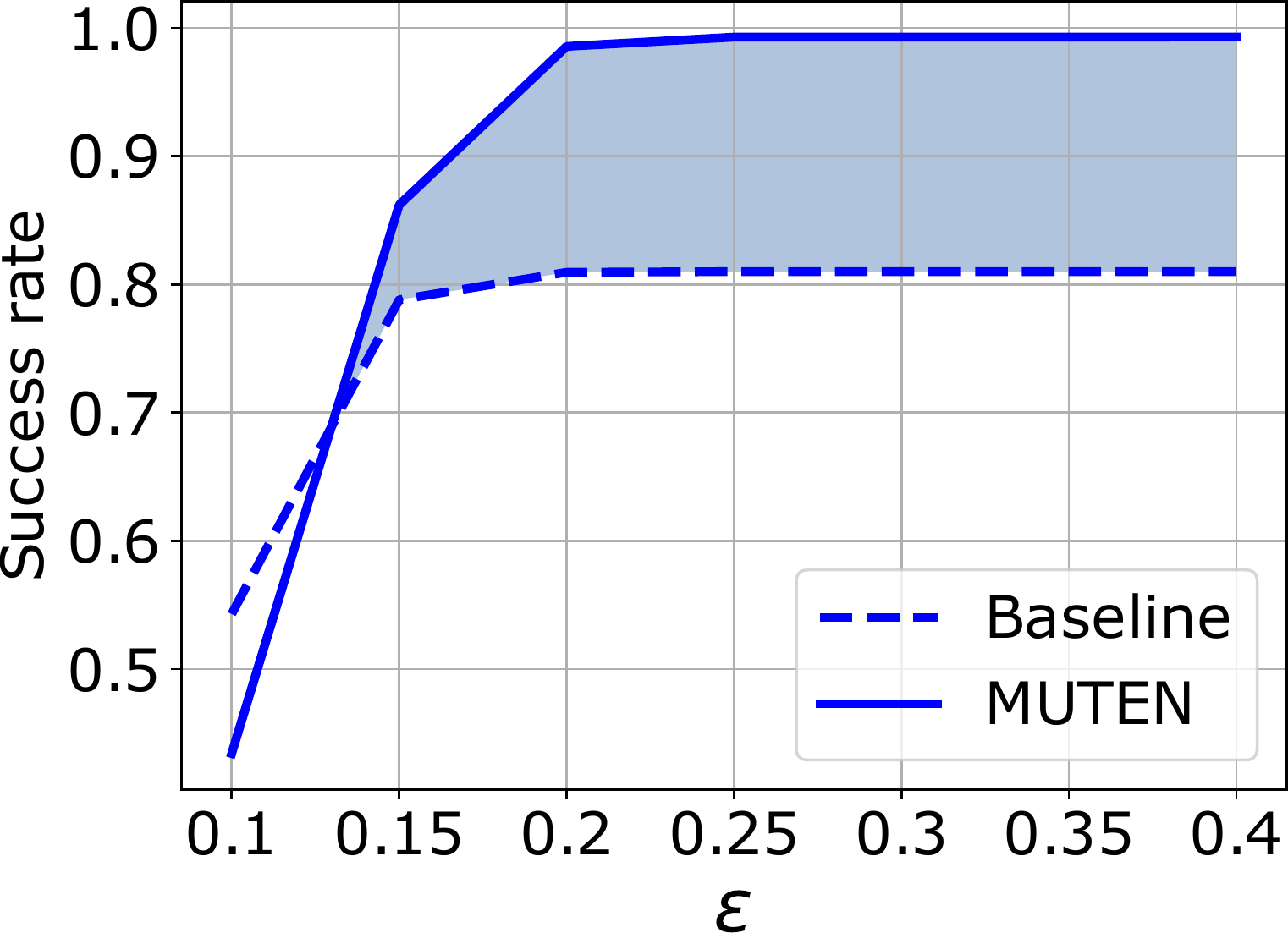}%
    }
    \subfigure[SVHN, BIM]{
    \includegraphics[scale=0.26]{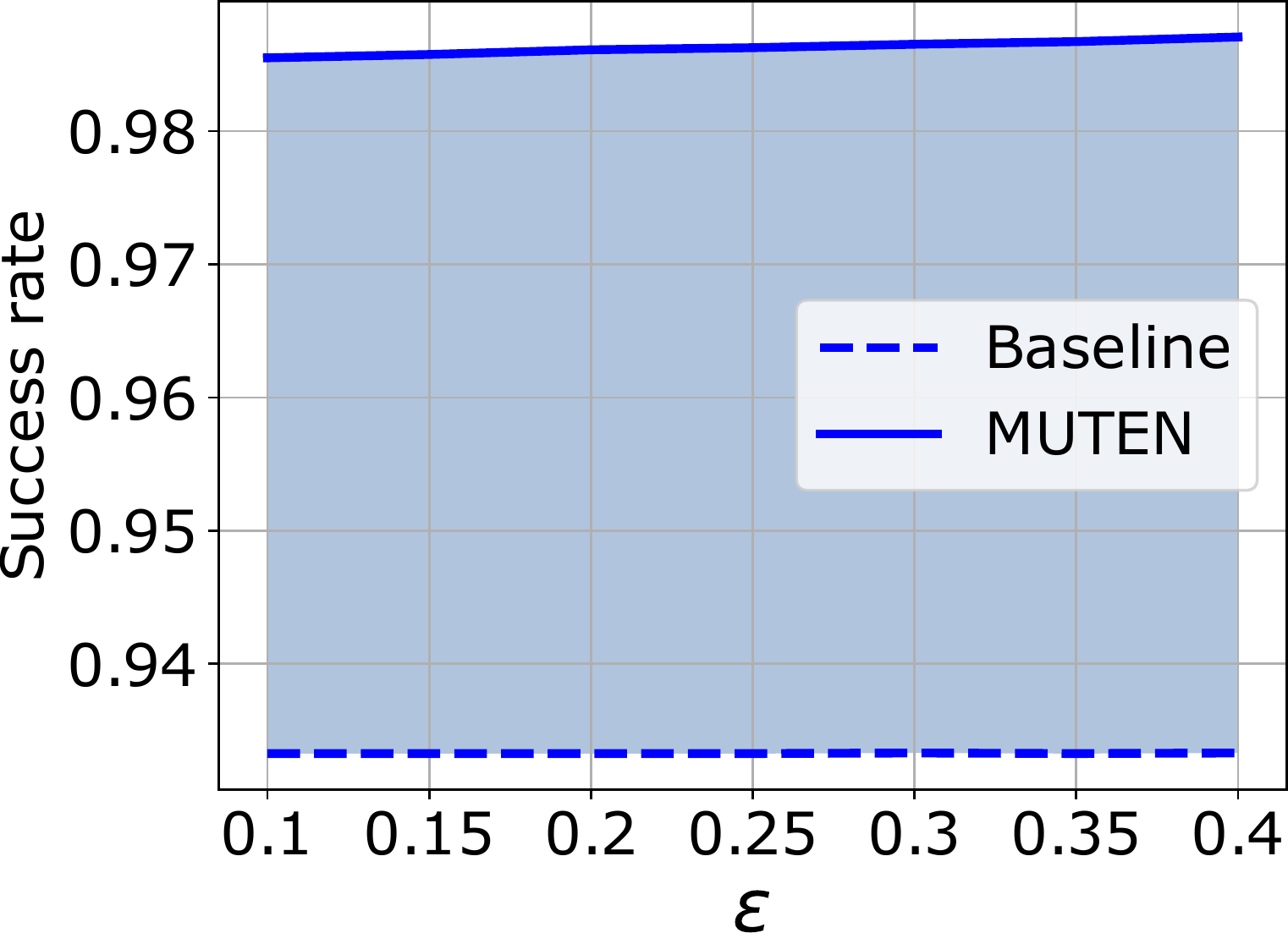}%
    }
    \subfigure[ResNet20V1, BIM]{
    \includegraphics[scale=0.26]{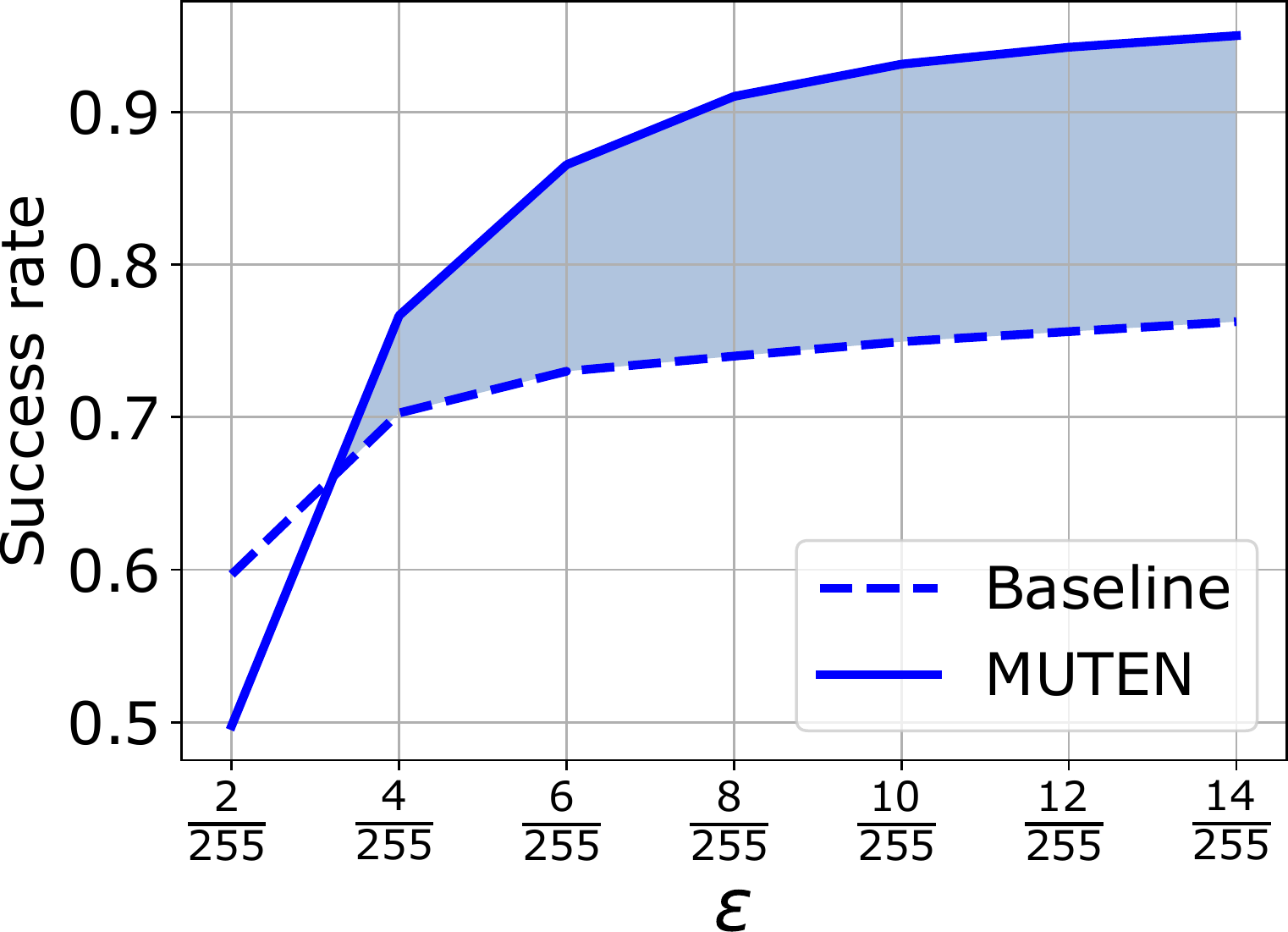}%
    }
    \subfigure[VGG16, BIM]{
    \includegraphics[scale=0.26]{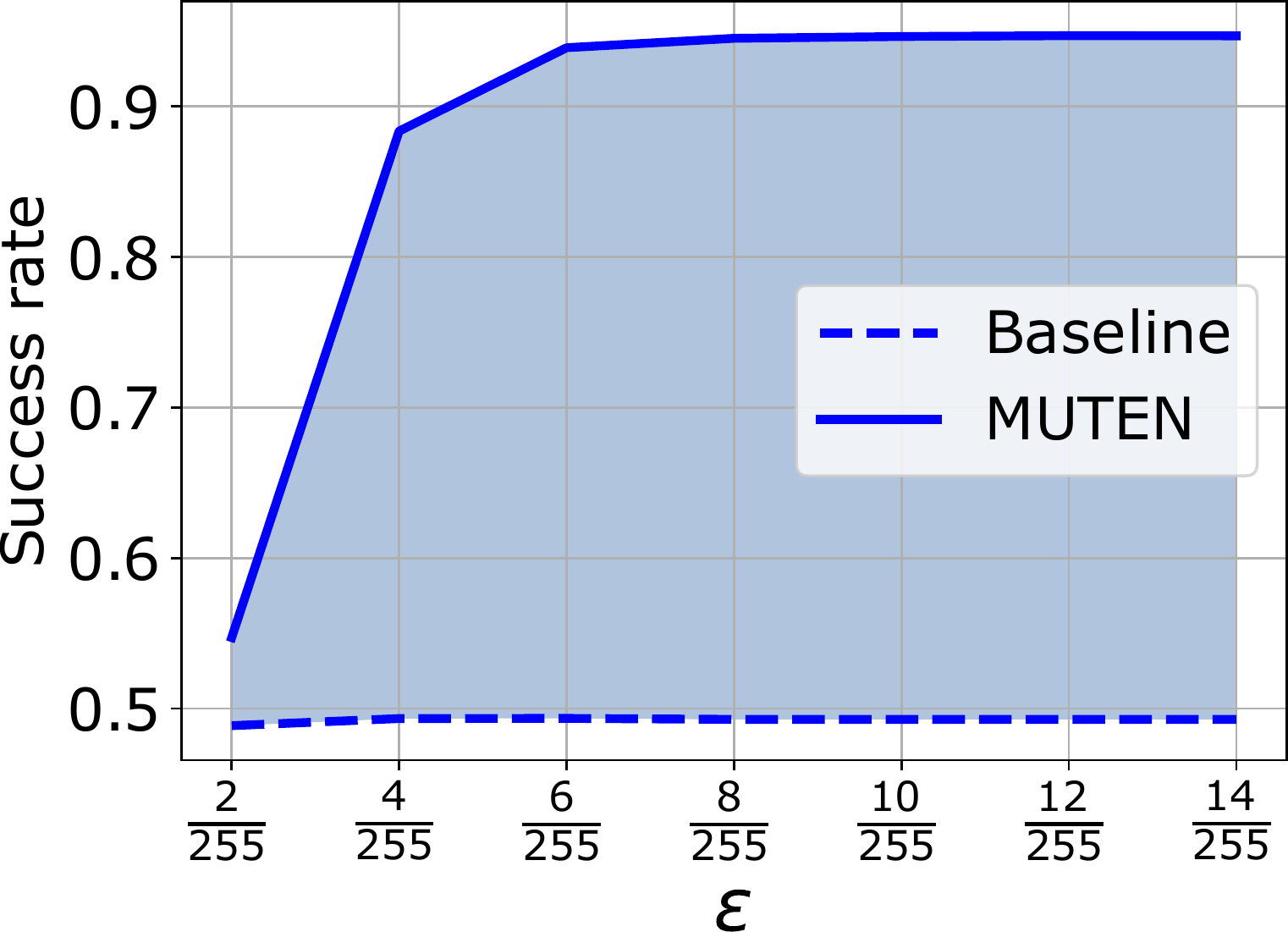}%
    }\\
    \vspace{-1.5mm}
    \subfigure[MNIST, PGD]{
    \includegraphics[scale=0.26]{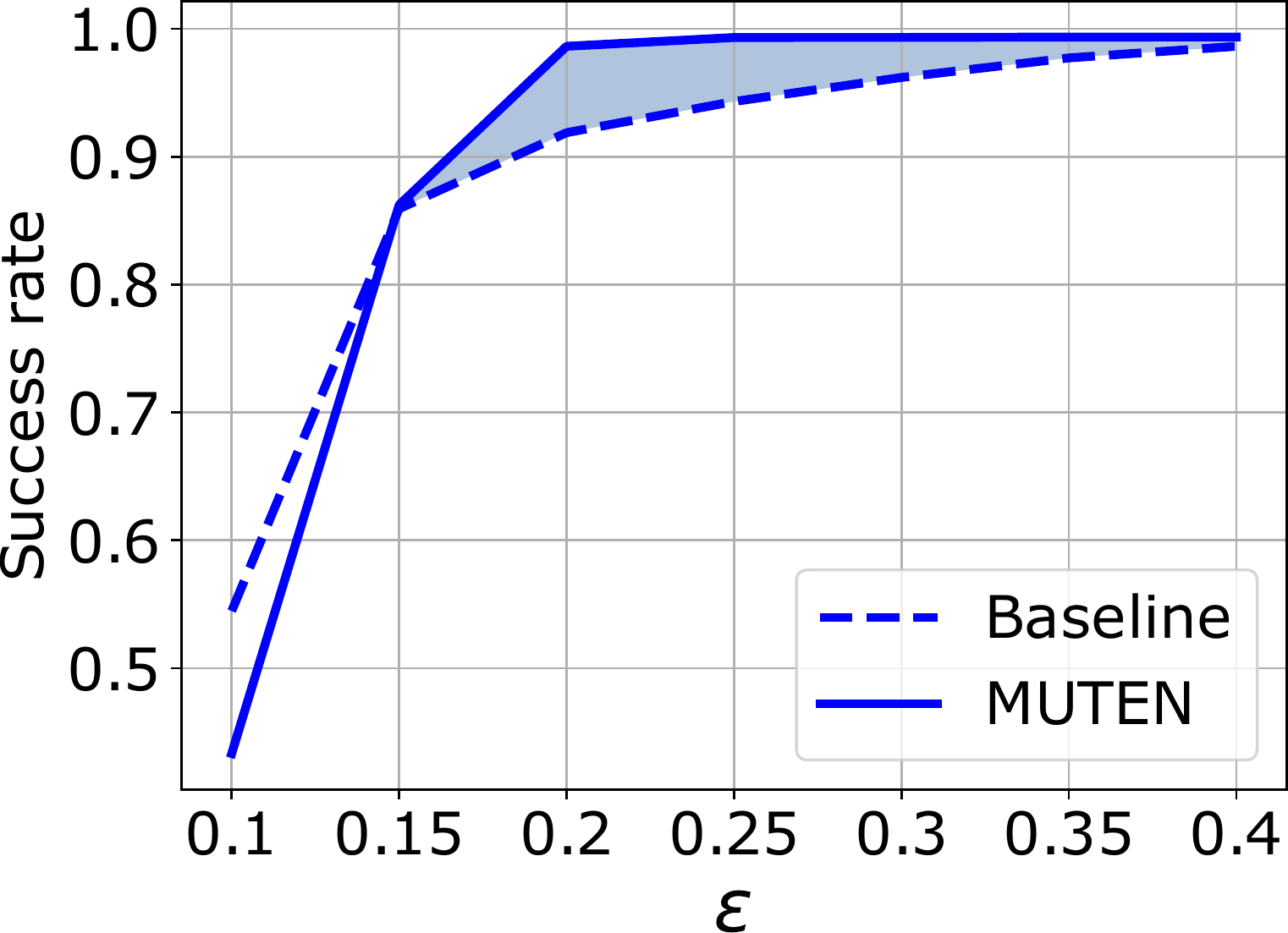}%
    }
    \subfigure[SVHN, PGD]{
    \includegraphics[scale=0.26]{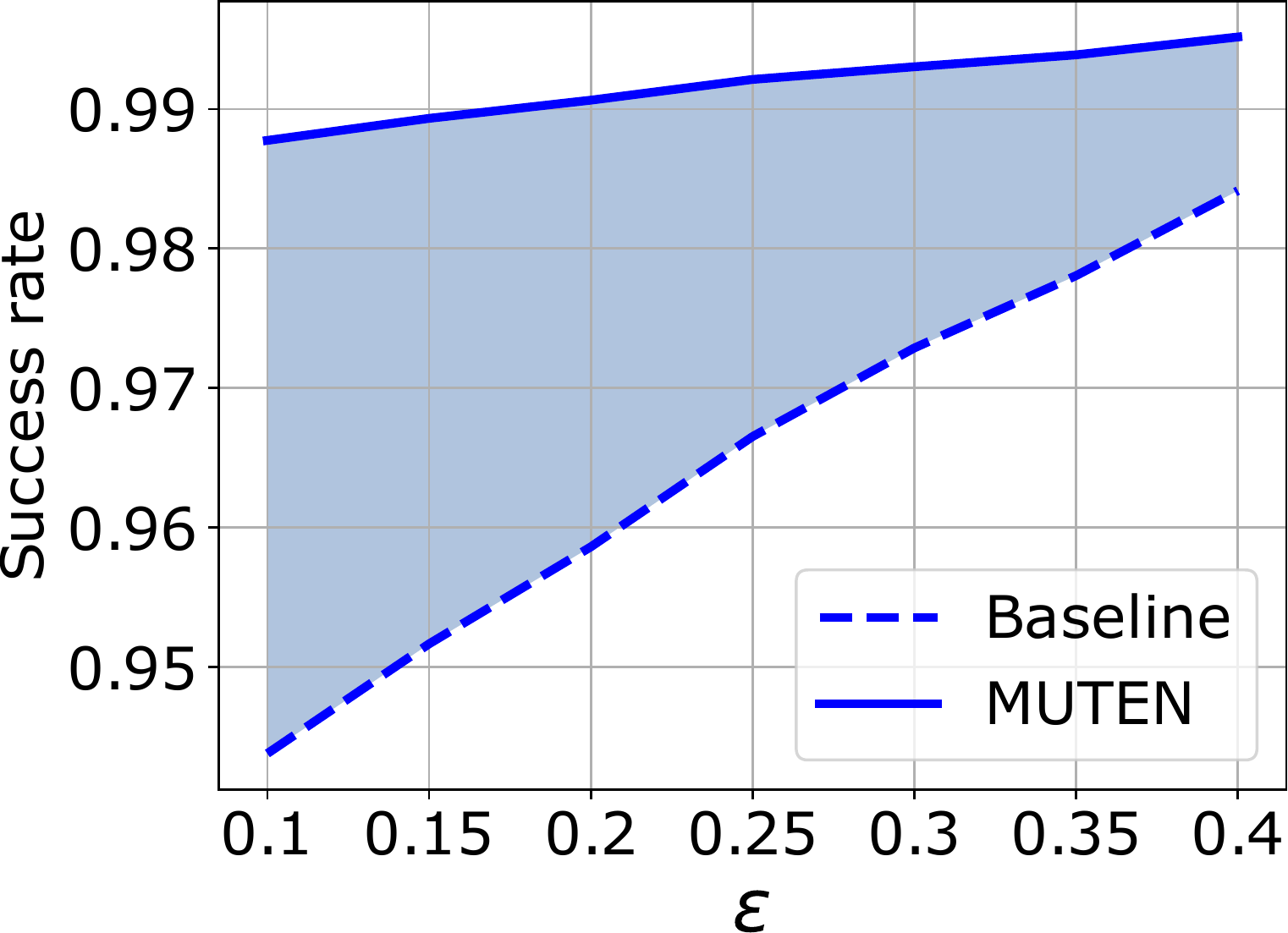}%
    }
    \subfigure[ResNet20V1, PGD]{
    \includegraphics[scale=0.26]{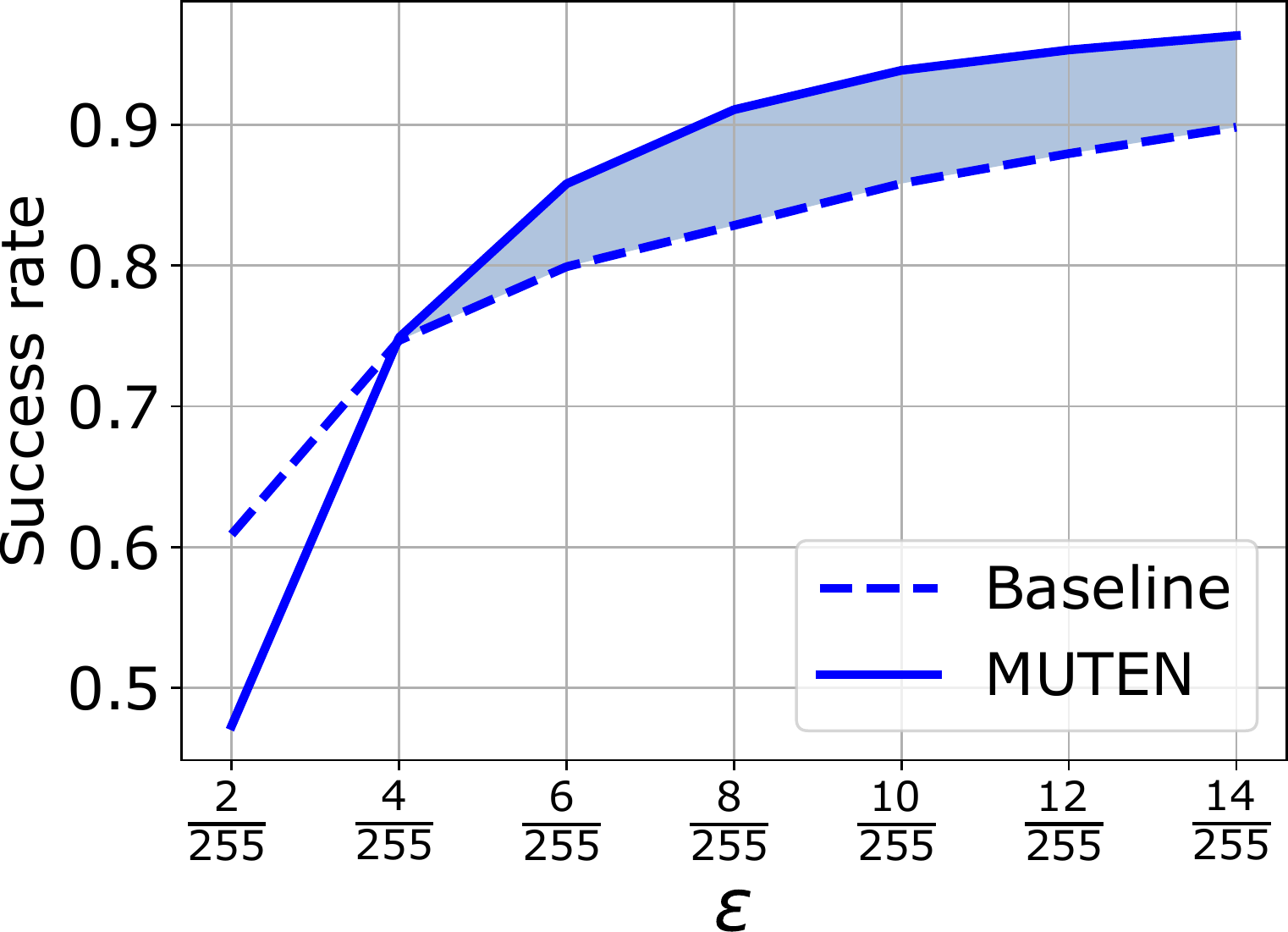}%
    }
    \subfigure[VGG16, PGD]{
    \includegraphics[scale=0.26]{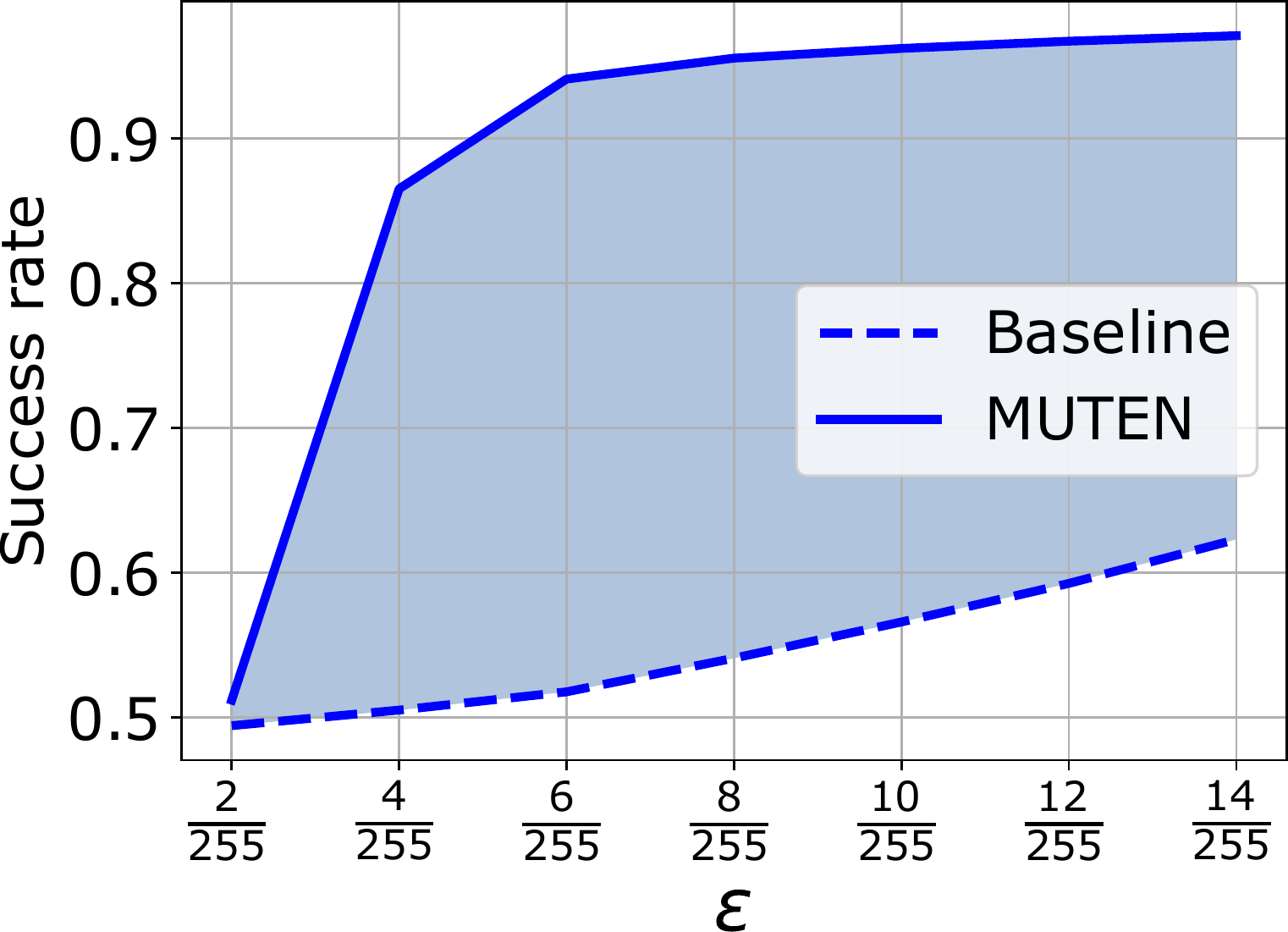}%
    }\\
    \vspace{-1.5mm}
    \subfigure[MNIST, C\&W]{
    \includegraphics[scale=0.26]{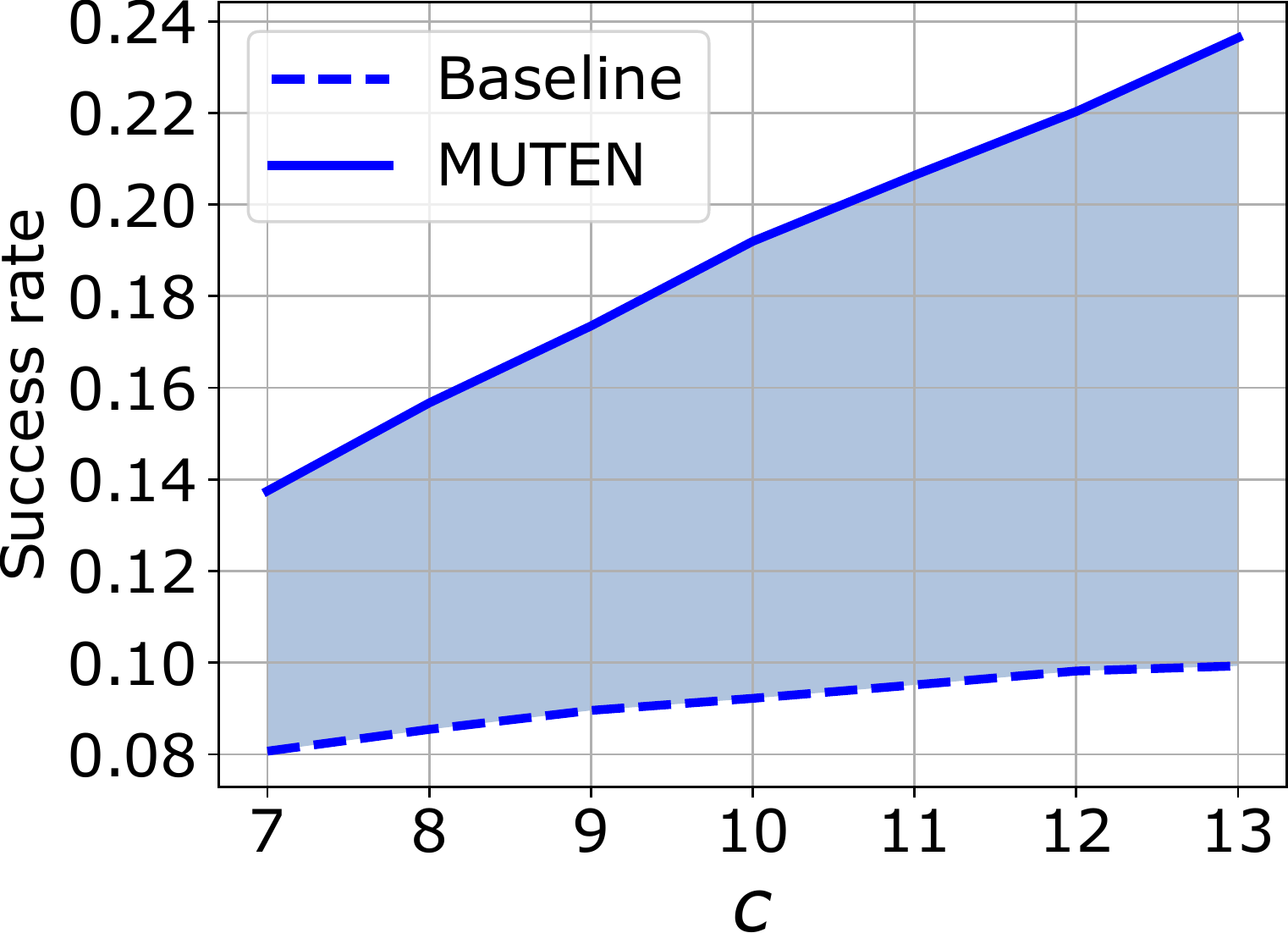}%
    }
    \subfigure[SVHN, C\&W]{
    \includegraphics[scale=0.26]{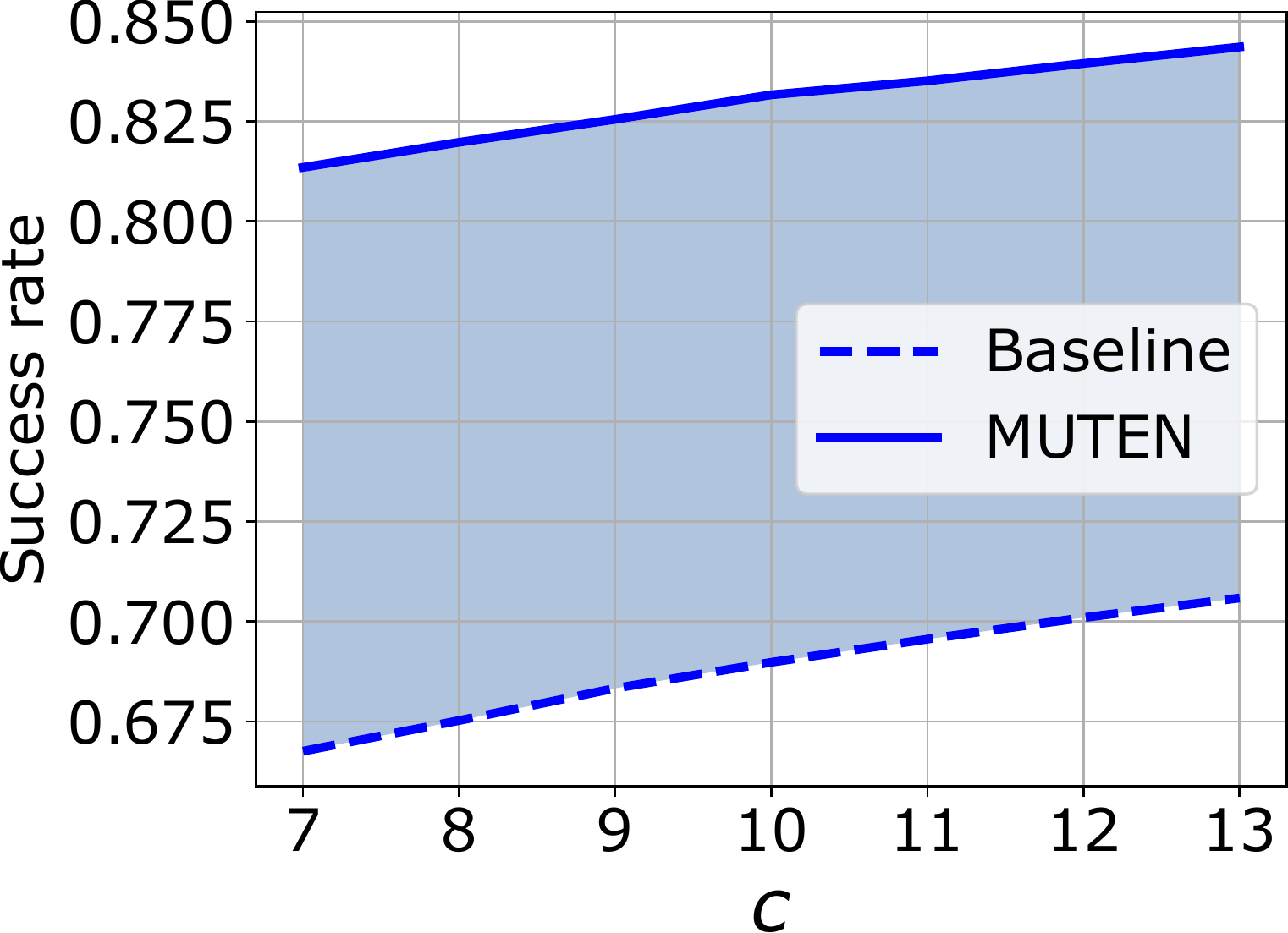}%
    }
    \subfigure[ResNet20V1, C\&W]{
    \includegraphics[scale=0.26]{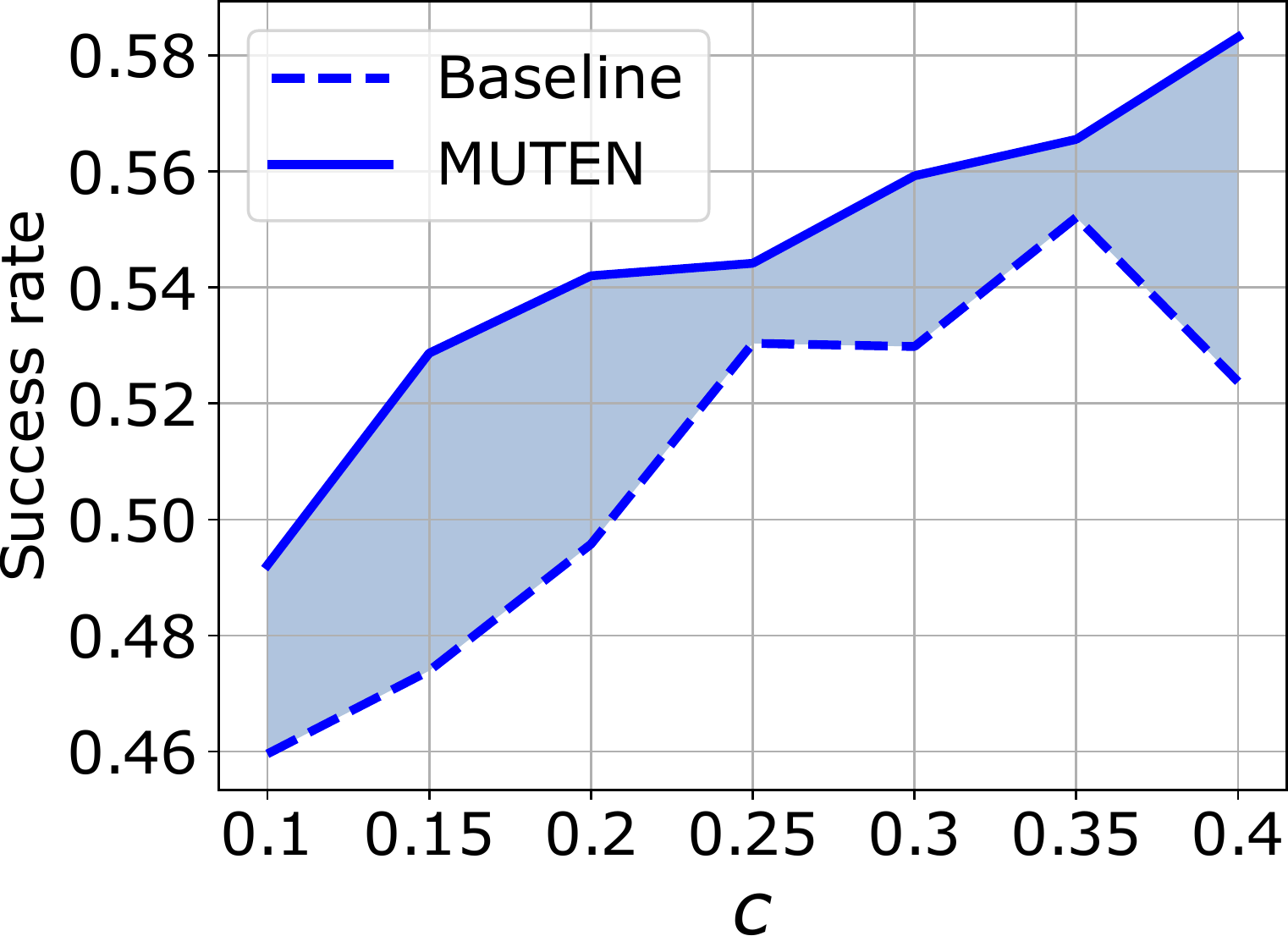}
    }
    \subfigure[VGG16, C\&W]{
    \includegraphics[scale=0.26]{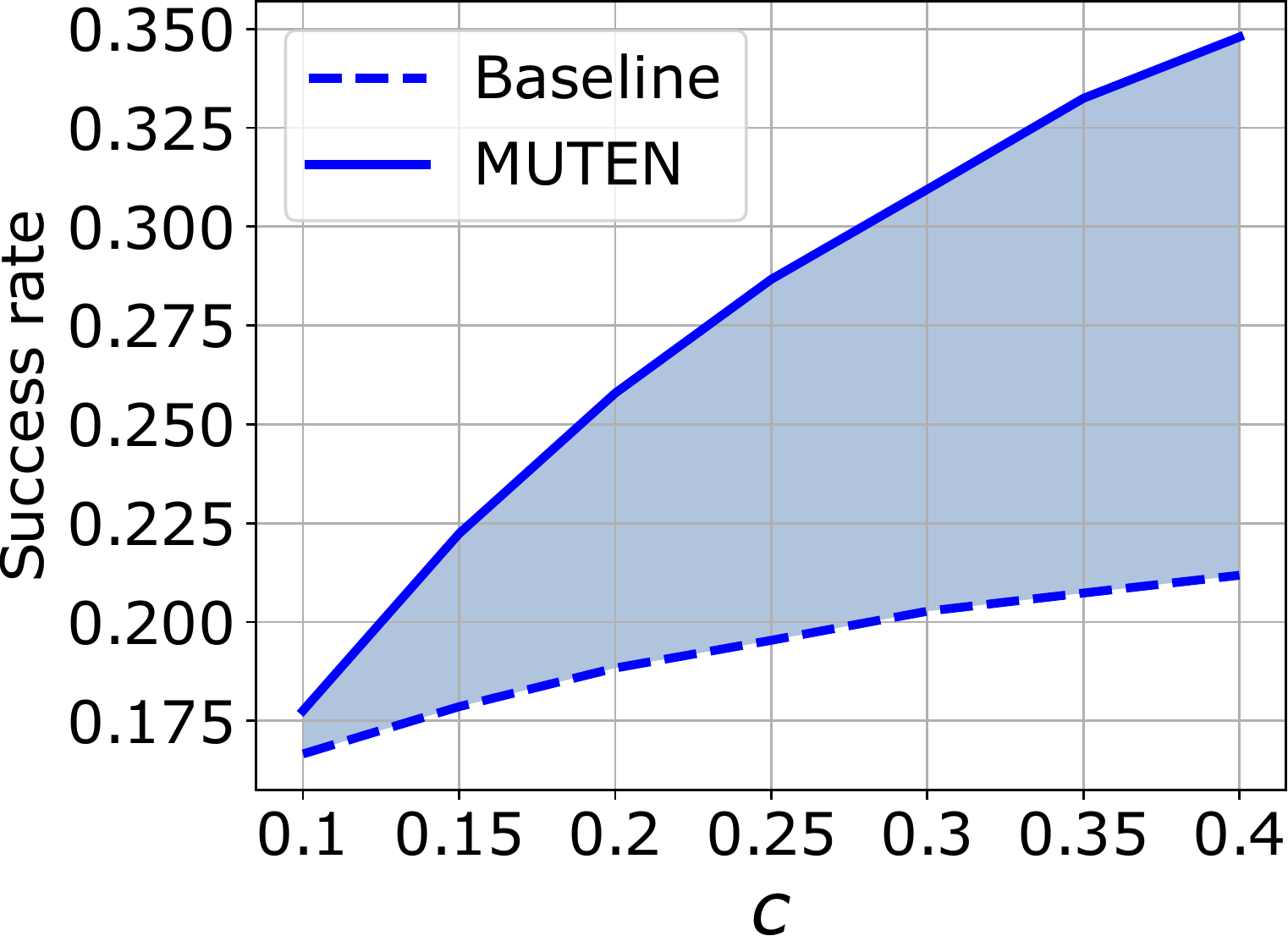}%
    }
    \caption{Success rate VS. attack configuration. The shade area indicates where MUTEN outperforms the baseline.}
    \label{fig:exp_1}
\end{figure*}

\paragraph{Effectiveness} Firstly, we evaluate the effectiveness of MUTEN measured as the success rate of the applied attacks. The success rate of the four attacks applied to the original model is taken as the baseline. Figure \ref{fig:exp_1} shows the success rate of MUTEN with different attack configurations. As the success rate converges mostly when the ensemble includes 5 mutants (see next section), we only present the result with this number. Overall, compared with the four baselines, MUTEN achieves a higher success rate, especially as the maximum perturbation size increases. For example, in the case of PGD with VGG16, the success rate by MUTEN reaches 0.97, while the baseline increases it by 0.62 with the maximum perturbation size. When the threshold is very small (for instance, $\epsilon<0.2$ for MNIST and $\epsilon<\frac{8}{255}$ for ResNet20V1 by FGSM), the success rate achieved by our approach is lower than or keeps similar to the baseline. The reason is that the ensemble is an average over multiple models and so as the gradient, given a very small perturbation, the attack requires more iterations to converge, which is demonstrated by FGSM and BIM (which is an iterative version of FGSM). 

\begin{figure}
\centering
    \subfigure[FGSM]{
    \includegraphics[scale=0.26]{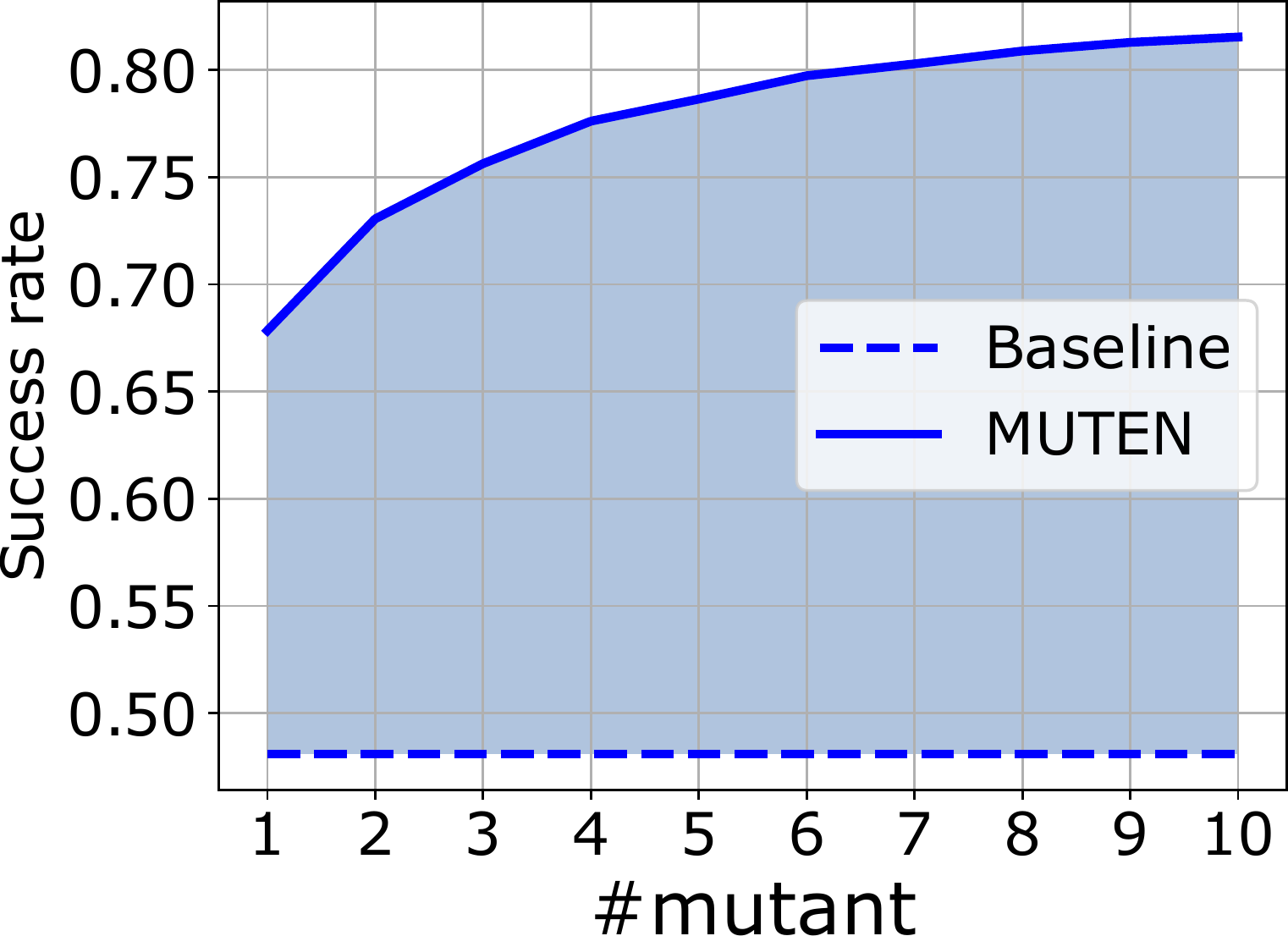}%
    }
    \subfigure[BIM]{
    \includegraphics[scale=0.26]{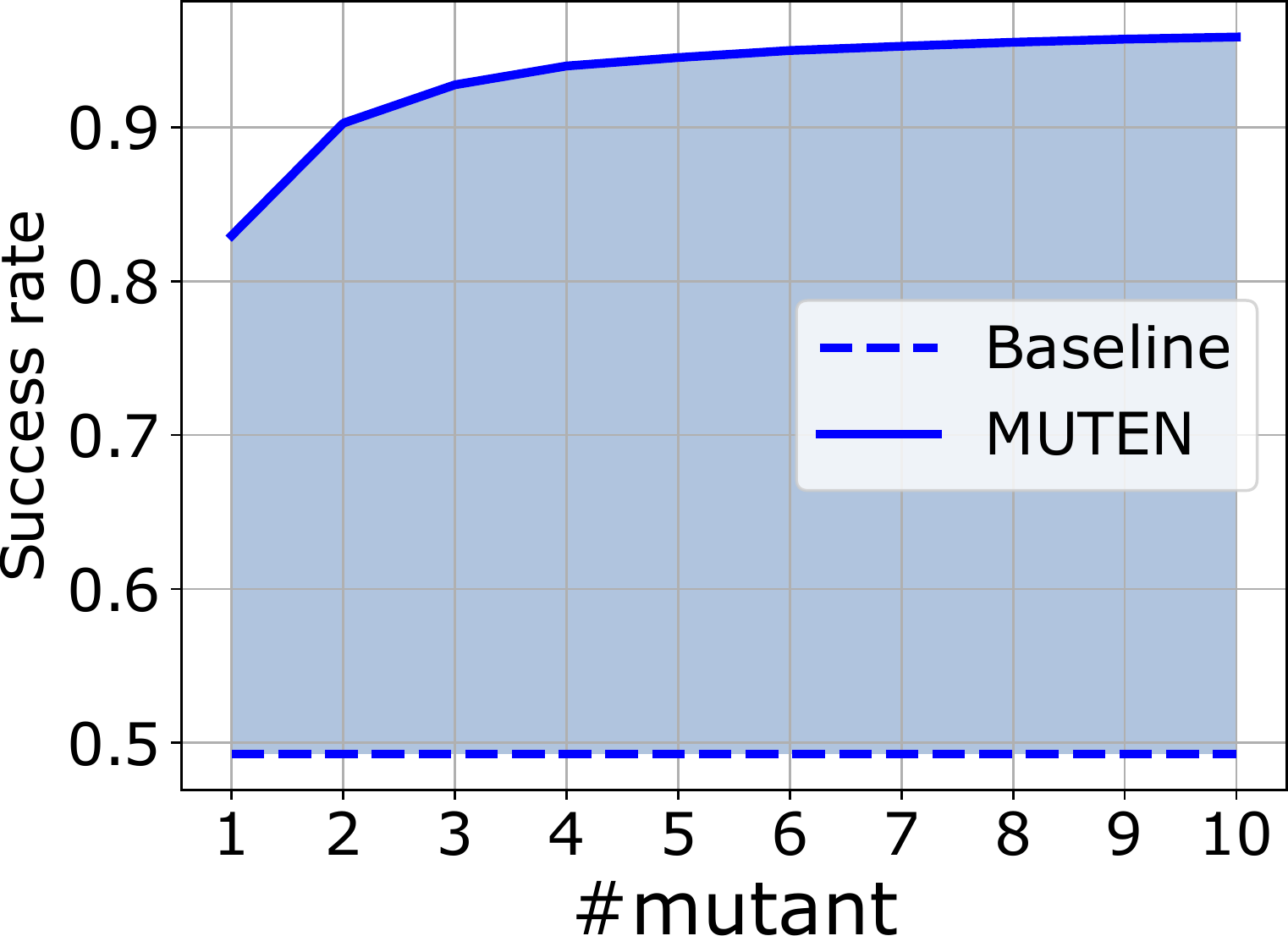}%
    }\\
    \subfigure[PGD]{
    \includegraphics[scale=0.26]{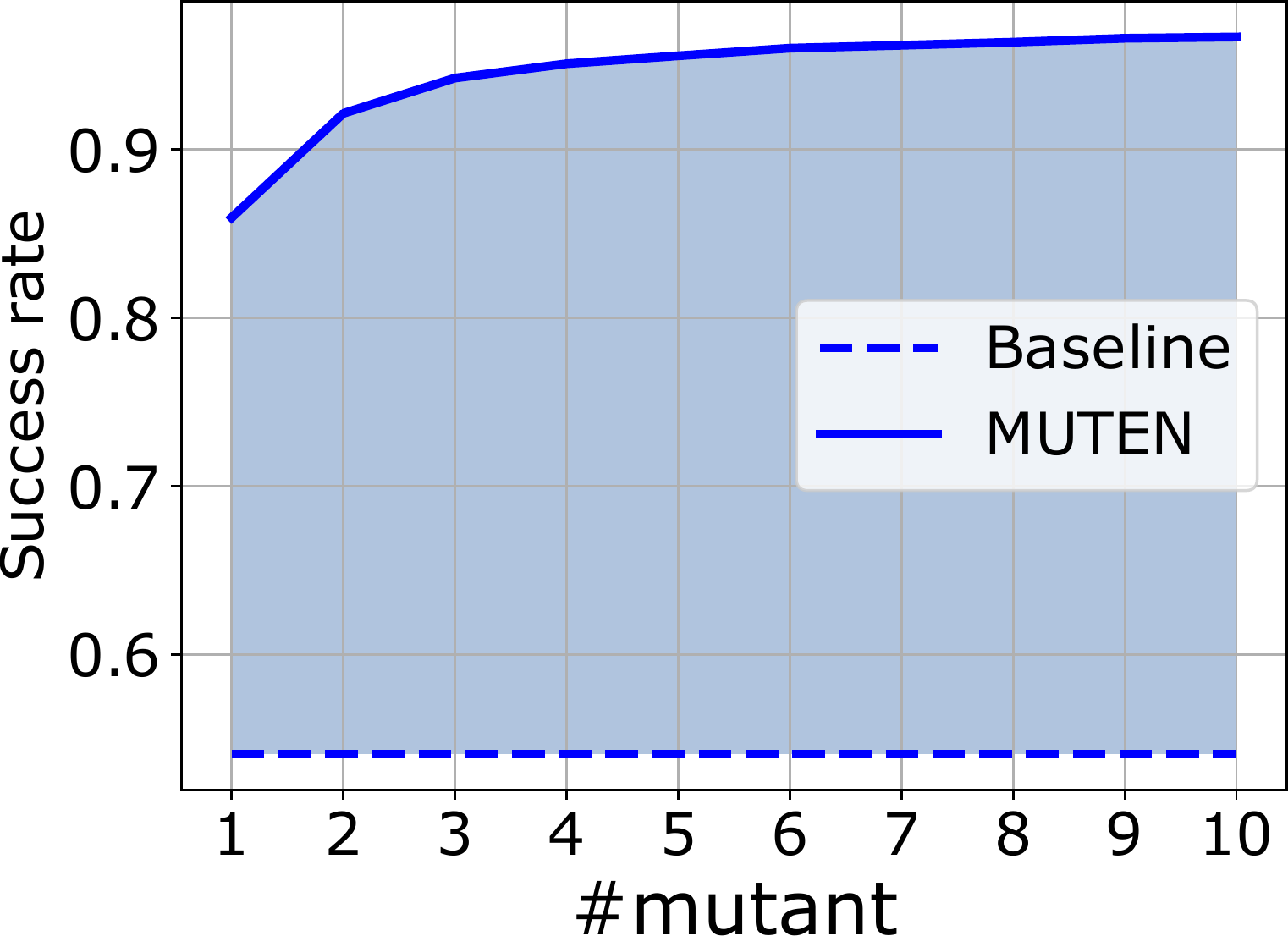}%
    }
    \subfigure[C\&W]{
    \includegraphics[scale=0.26]{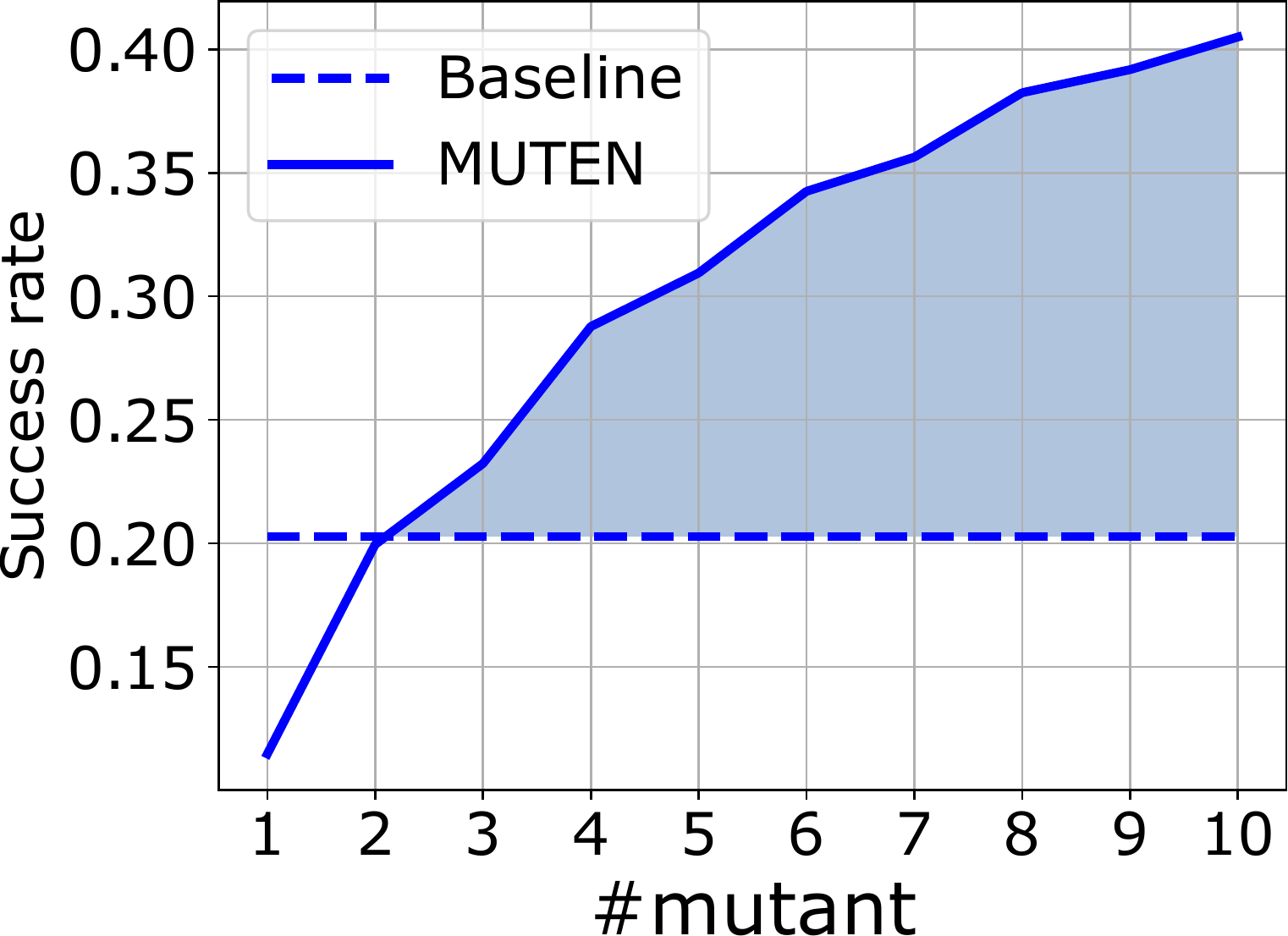}%
    }
    \caption{Success rate VS. number of mutants. The shade area indicates where MUTEN outperforms the baseline. Dataset: CIFAR10; Model: VGG16.}
    \label{fig:exp_2}
\end{figure}

\begin{figure}
\centering
    \subfigure[FGSM]{
    \includegraphics[scale=0.26]{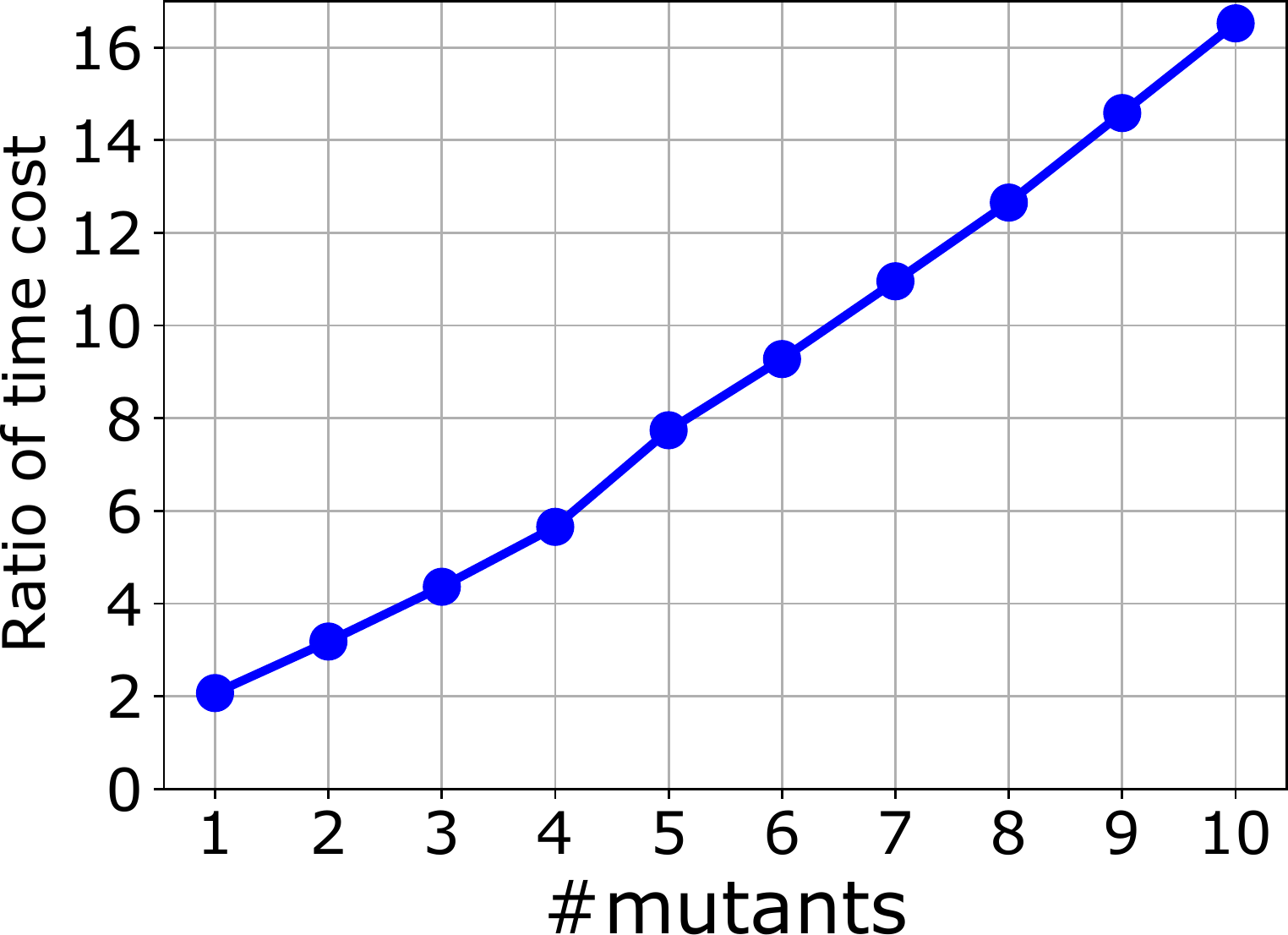}%
    }
    \subfigure[BIM]{
    \includegraphics[scale=0.26]{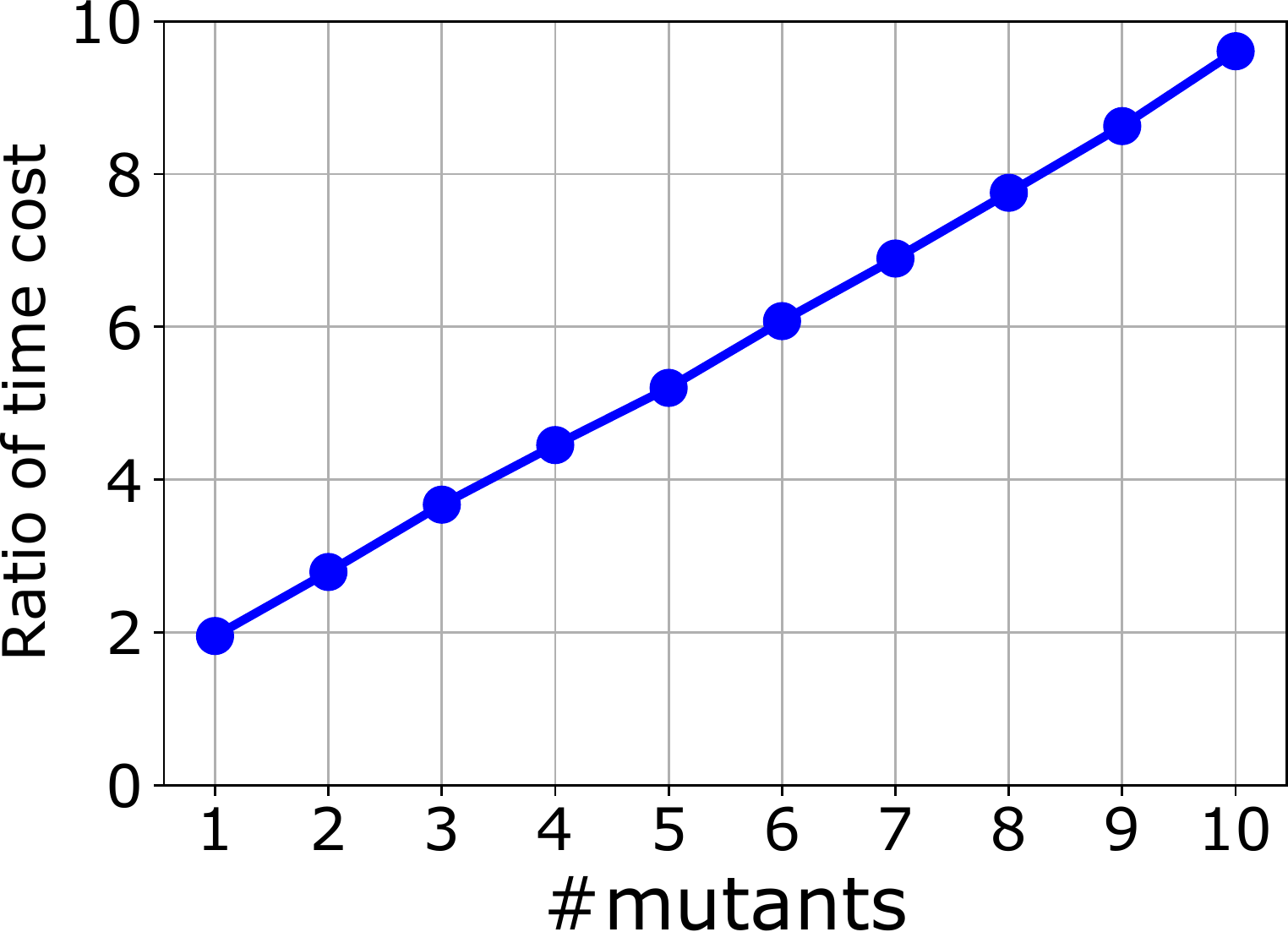}%
    }\\
    \subfigure[PGD]{
    \includegraphics[scale=0.26]{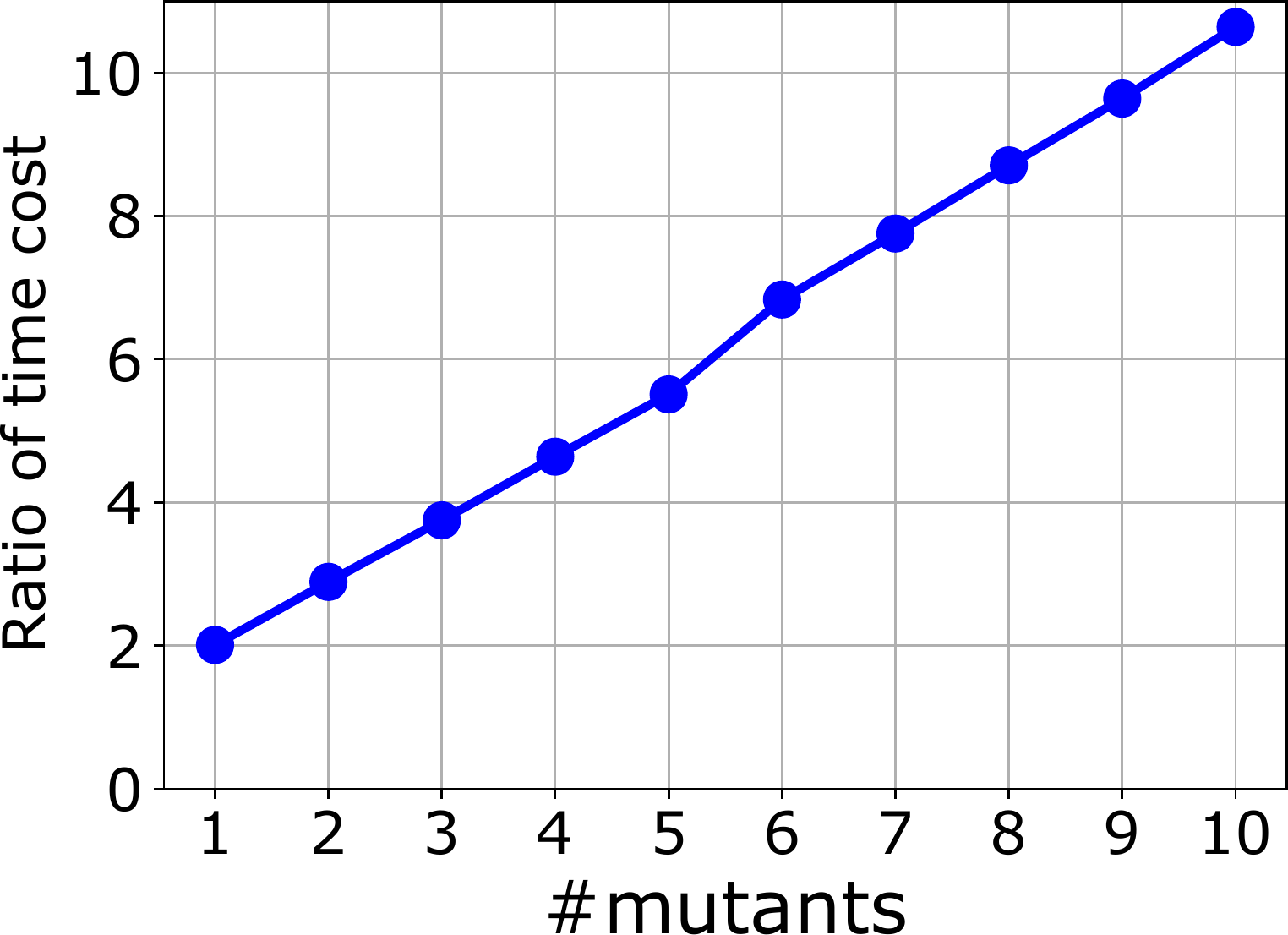}%
    }
    \subfigure[C\&W]{
    \includegraphics[scale=0.26]{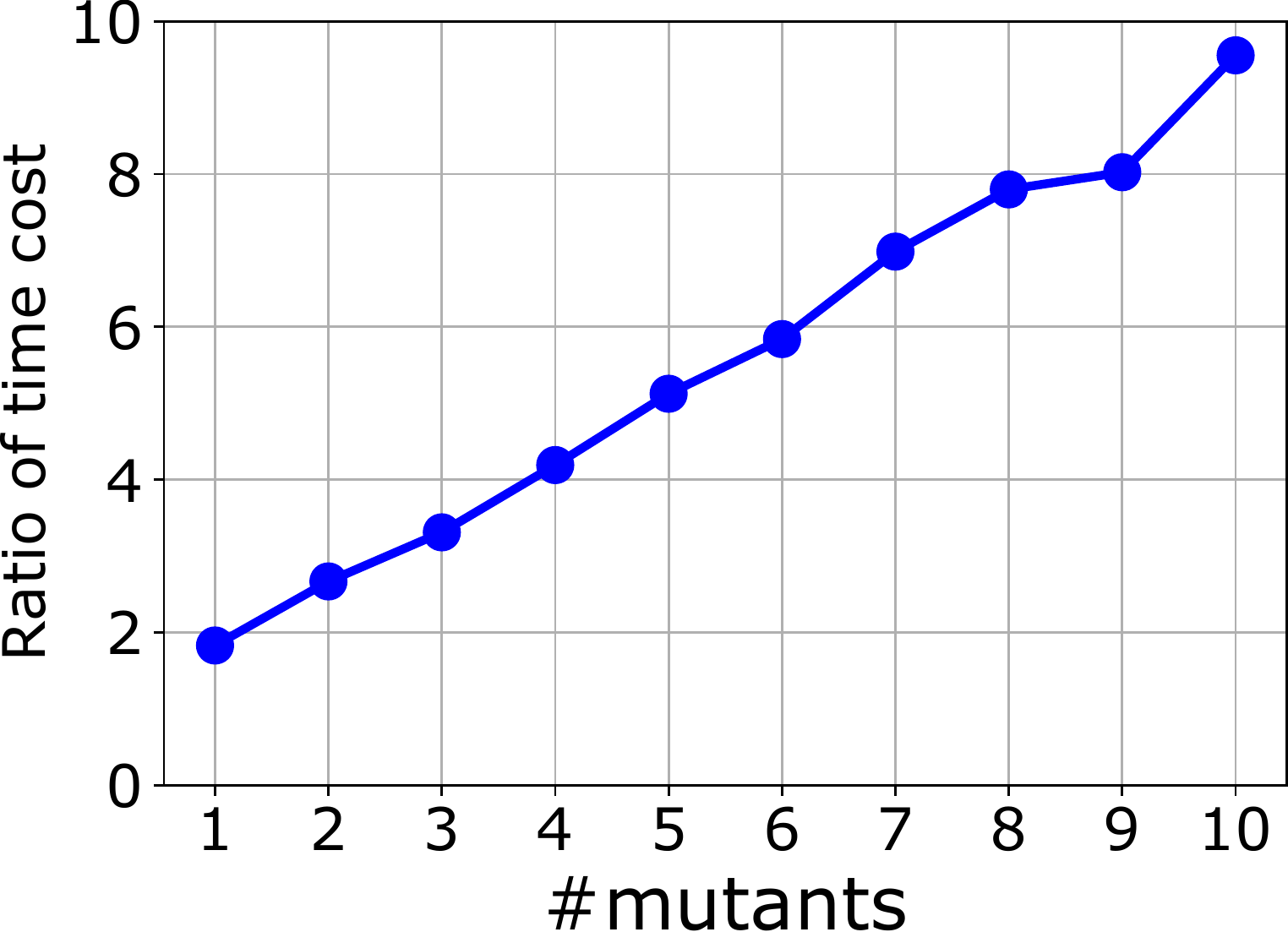}%
    }
    \caption{Ratio of time cost VS. number of mutants. Dataset: CIFAR10; Model: VGG16.}
    \label{fig:exp_2_time}
\end{figure}

\paragraph{Impact of the number of mutants} Secondly, we investigate how the number of mutants in the ensemble model impacts the effectiveness of MUTEN. We use the commonly used configurations ($\epsilon=\frac{8}{255}$, $c=0.3$) for the attacks and consider a number of mutants ranging from 1 to 10. Figure \ref{fig:exp_2} shows the results for CIFAR10 with VGG16. In general, the effectiveness of MUTEN increases as more mutant is integrated but tends to saturate quickly in the cases of FGSM, BIM and PGD. Concerning the improvement of success rate, it increases from 0.20 to 0.33 in FGSM, 0.34 to 0.47 in BIM, 0.32 to 0.43 in PGD, and -0.09 to 0.21 in C\&W. In the case of FGSM, IFGSM, and PGD, a very high improvement of success rate can be reached with only 1 mutant, which is only the double-time of attacking the original model. Adding more mutants will increase more the success rate but with a slower growth rate. In the case of the strongest attack, C\&W, the success rate is lower than the baseline when the ensemble includes 1 mutant, which also happens to the other models. The reason is that for C\&W, when the parameter $c$ is very small, the gradient loss has a small contribution to the loss function used by the attack algorithm (see Section \ref{sec:background-attack}). By contrast, letting the gradient loss be more important by increasing $c$, the success rate boosts quickly. Thus, in this case, to increase the success rate of C\&W, one can either adjust $c$ to be greater or include more mutants. Note that, attacking an ensemble requires multiple times due to that the attack computes the average gradient over each base model. As a result, the time cost is linearly related to the number of models in the ensemble. 
Figure \ref{fig:exp_2_time} shows the ratio of time cost of attacking an ensemble to a single model. 
Therefore, in practical applications, one has to consider the trade-off between a high success rate and a small number of mutants.

\begin{figure}[!ht]
\centering
    \subfigure[FGSM]{
    \includegraphics[scale=0.26]{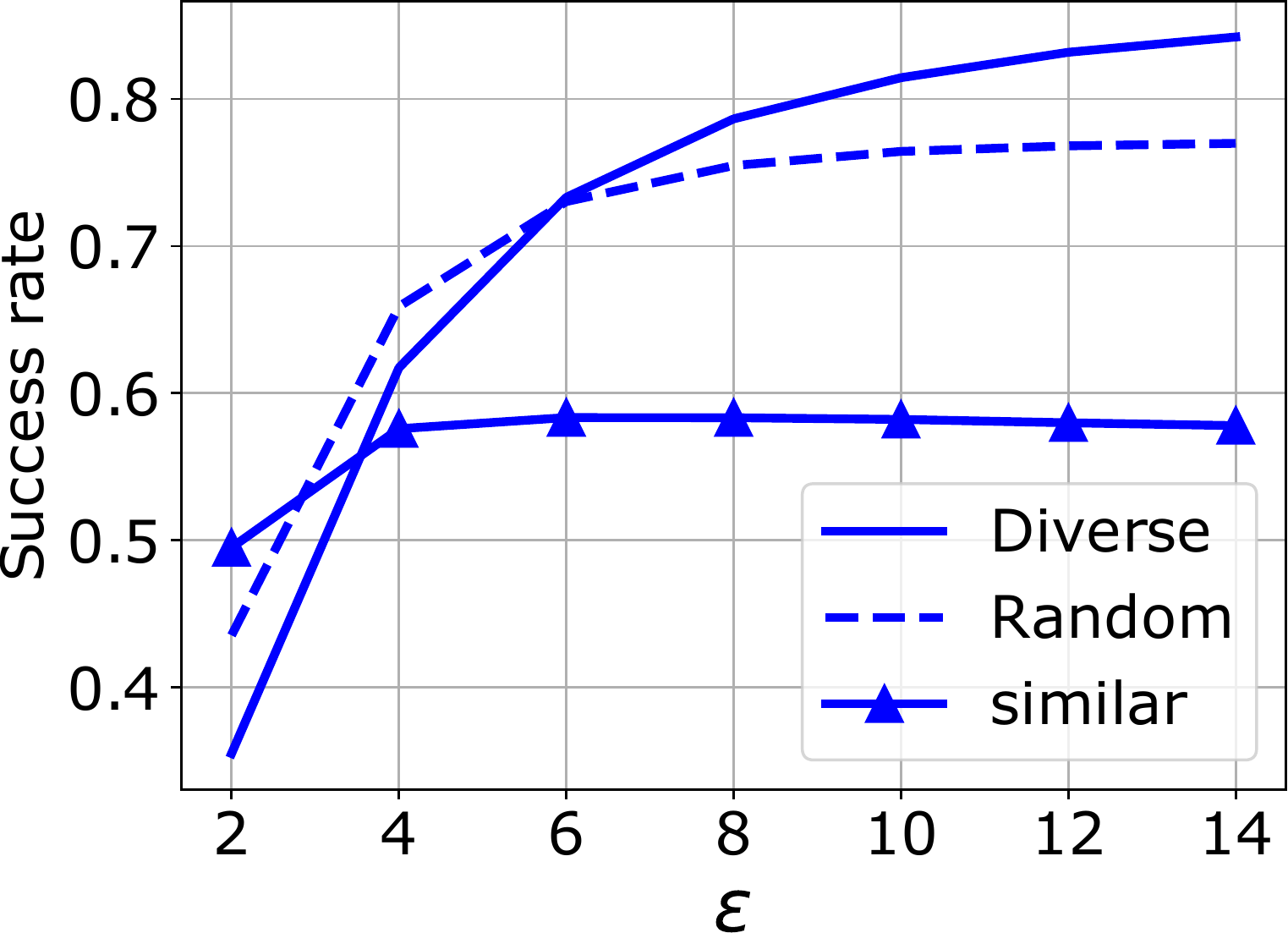}%
    }
    \subfigure[BIM]{
    \includegraphics[scale=0.26]{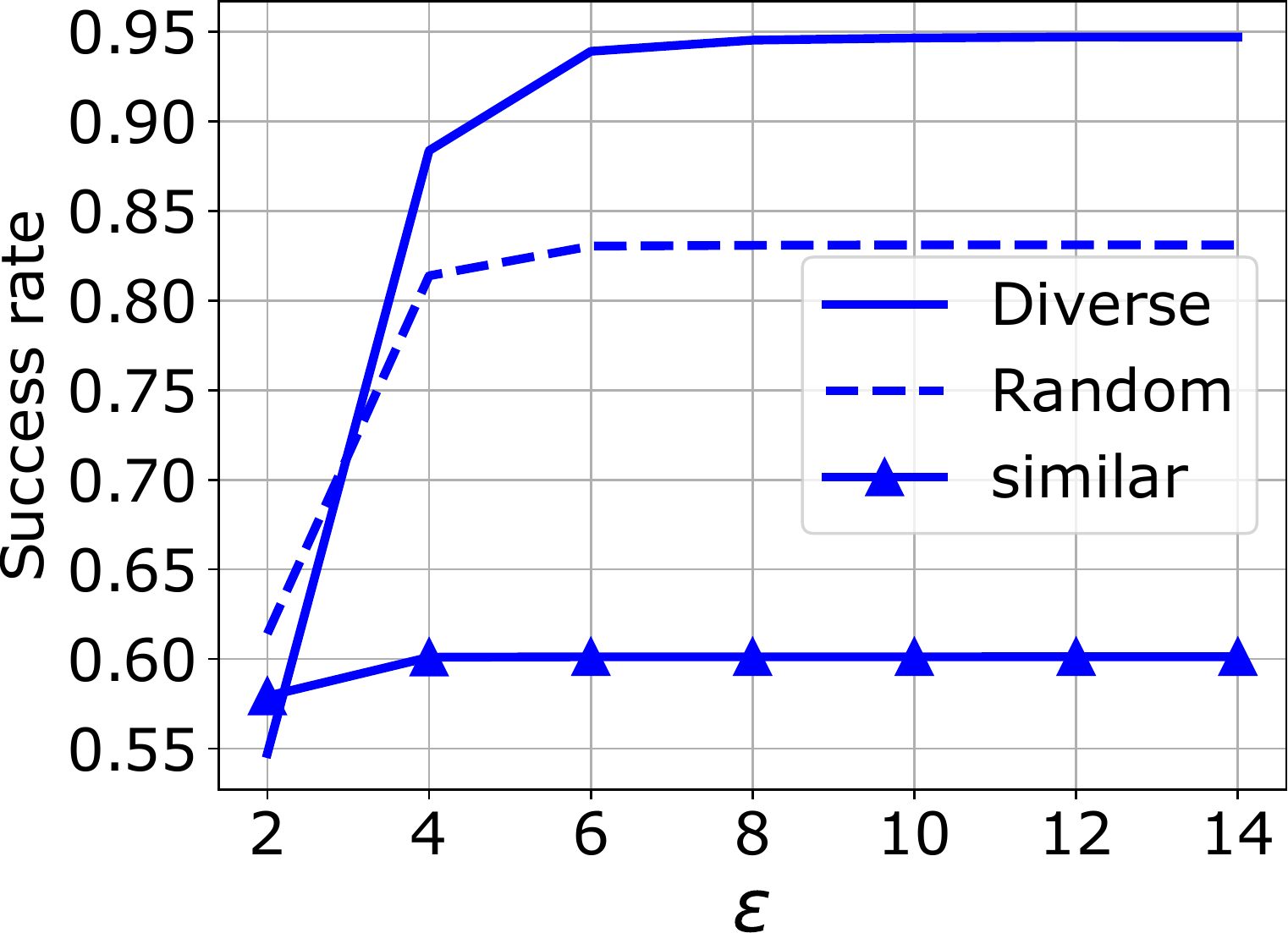}%
    }\\
    \subfigure[PGD]{
    \includegraphics[scale=0.26]{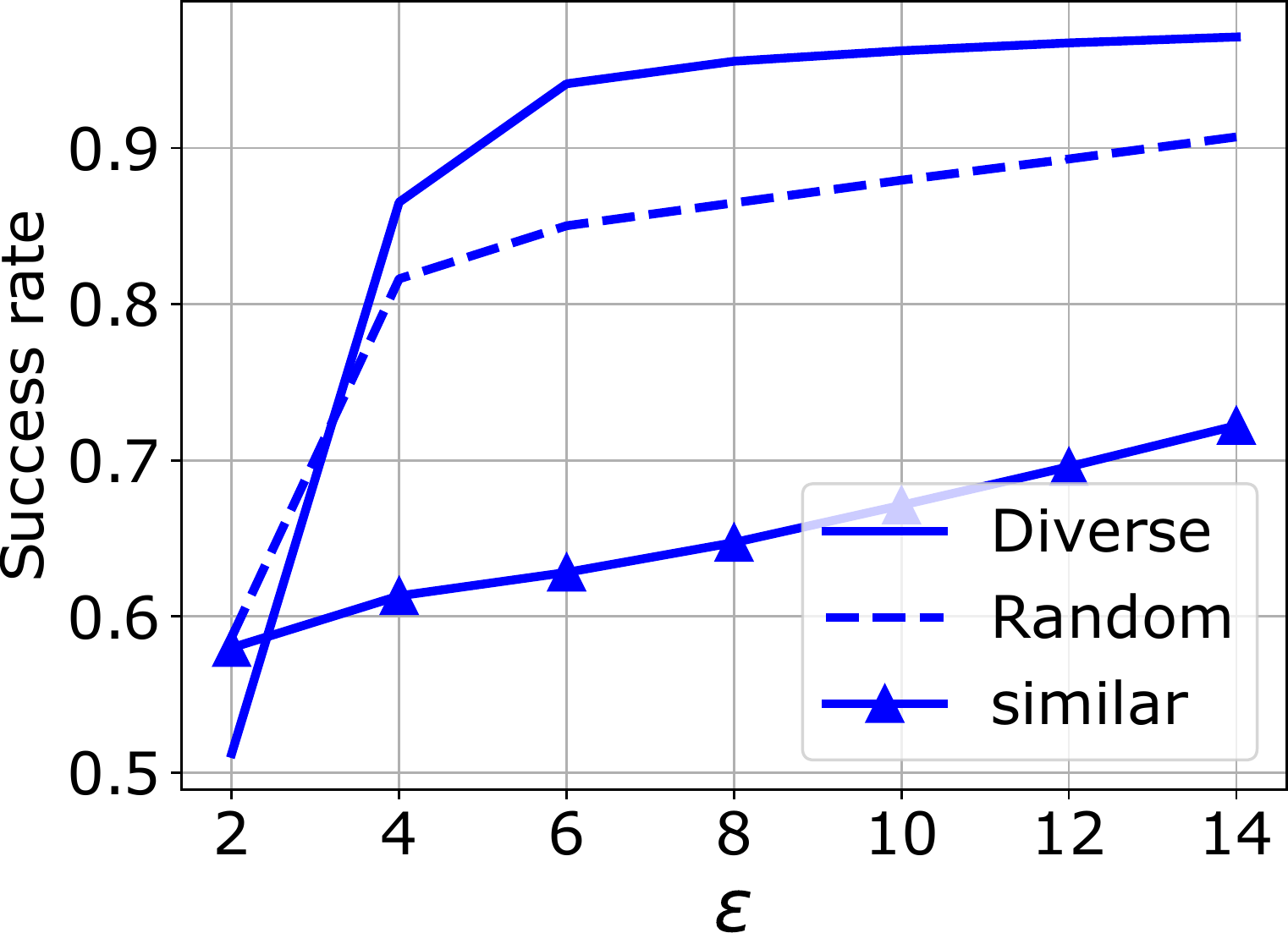}%
    }
    \subfigure[C\&W]{
    \includegraphics[scale=0.26]{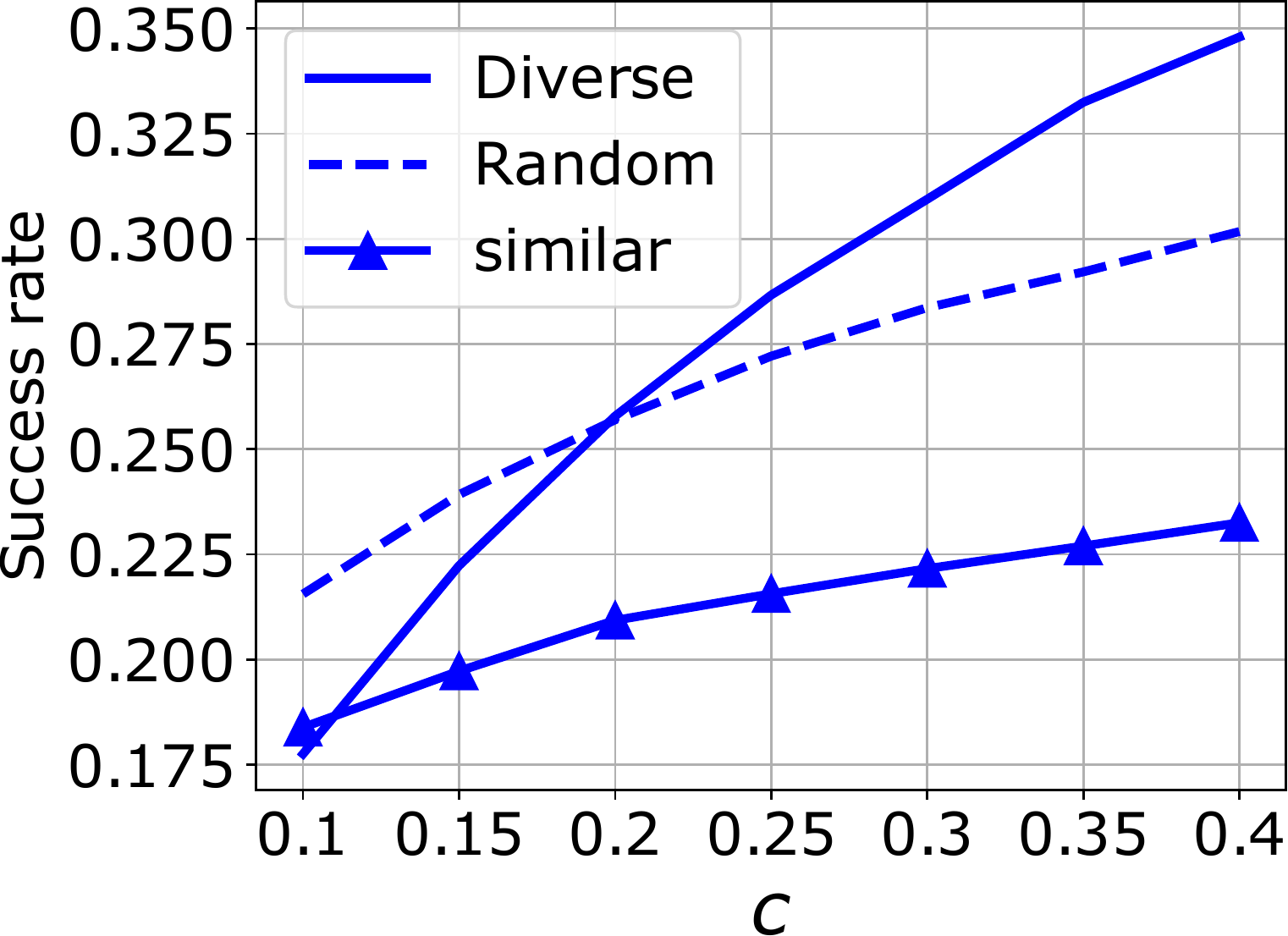}%
    }
    \caption{Success rate VS. type of mutants. Each ensemble model includes 5 mutants. The shade area indicates where MUTEN outperforms the baseline. Dataset: CIFAR10; Model: VGG16.}
    \label{fig:exp_3}
\end{figure}

\paragraph{Diversity of mutants} Thirdly, we investigate if the diversity contributes to MUTEN. We produce three types of mutants, diverse, random, and similar. The random mutants are generated by random selection, and similar mutants are created by limiting the test accuracy to be at least 95\% reserved for the original model. Figure \ref{fig:exp_3} shows the result of CIFAR10 with model VGG16. In general, the diverse ensemble performs the best, and the random ensemble works better than the similar one. In the case of FGSM, BIM, and PGD, when the perturbation is small (for instance, smaller than $\frac{4}{255}$), the difference of using three types of mutants is not big. By increasing the number of iterations of the greedy algorithm, the mutants can be more diverse, and the difference between random and diverse ensembles will be greater. 

\subsection{Discussion}
\paragraph{Extensions of MUTEN} Overall, using the diverse mutant-based ensemble is helpful to improve the effectiveness of craft adversarial examples. As it is easy and simple to generate a large number of mutants, MUTEN can be taken as a plug-in module in white-box attacks. Besides, since the number of mutants linearly relates to the efficiency, the number can be adjusted to balance between the time cost and the desired number of adversarial examples.

\paragraph{Threats to validity} The internal threat comes from our implementation. To counter this, we use some popular libraries as mentioned in the section of experiments, and carefully check the code. The external threat is the selected DNN models and datasets. We choose popular datasets with well-known models, and the model ranges from simple to complex. The construct threat is mainly about the method. As the diversity of mutants is important to obtain a good result, the metric to measure the similarity between mutants and the method to rank the diversity of mutants, and the ensemble strategy \cite{demir2016ensemble} could be changed to improve the performance.

\section{Conclusion}
We proposed MUTEN to effectively improve the gradient-based adversarial attacks. The main idea is to build an ensemble by diverse mutants to modify the gradient for the attacks to easier figure out the perturbation direction. Besides, since no extra training is required to produce a large number of mutants, MUTEN can be regarded as a simple plug-in module in white-box attacks. The experiments on different datasets and models have demonstrated that MUTEN performs promisingly to increase the success rate of four state-of-the-art gradient-based adversarial attacks with only a few mutants. Additionally, the diversity of mutants have been verified to be necessary to achieve effectiveness.

\clearpage
\bibliographystyle{named}
\bibliography{ijcai21}

\end{document}